\documentclass[preprint, 3p, 12pt,sort&compress]{elsarticle}
\usepackage{amssymb}
\usepackage{booktabs}
\usepackage{xcolor}
\usepackage{enumitem}
\usepackage{lipsum}
\usepackage[normalem]{ulem}
\usepackage{xcolor}
\usepackage[version=4]{mhchem}
\usepackage{amsmath, amssymb}
\usepackage{soul,graphicx, float}
\usepackage{placeins}
\usepackage[colorlinks=true, allcolors=blue]{hyperref}
\usepackage{graphicx}   
\usepackage{adjustbox}  
\usepackage{caption}
\usepackage[numbers]{natbib}
\usepackage{comment}
\usepackage{caption}
\usepackage{caption}
\usepackage{amsthm}
\usepackage{subcaption}
\usepackage{amsmath}
\usepackage{algorithm}
\usepackage{algorithmicx}
\usepackage{algpseudocode}
\usepackage{multirow}

\theoremstyle{definition}
\newtheorem*{remark}{Remark}
\newtheorem*{definition}{Definition}
\newtheorem*{theorem}{Theorem}
\newtheorem*{corollary}{Corollary}
\newtheorem*{lemma}{Lemma}
\newtheorem*{assumption}{Assumption}
\newtheorem*{proposition}{Proposition}

\newcommand{\alr}[1]{\textcolor{red}{#1}}

\journal{Elsevier}

\begin{document}

\begin{frontmatter}



\title{SHA-256 Infused Embedding–Driven Generative Modeling of High-Energy Molecules in Low-Data Regimes}


\author[mech]{Siddharth Verma}
\affiliation[mech]{Department of Mechanical Engineering, organization={Indian Institute of Technology Bombay, Powai},
            city={Mumbai},
            country={India}}
\author[mech,cminds]{Alankar Alankar \corref{cor}}
\affiliation[cminds]{Center for Machine Intelligence and Data Science (CMInDS), organization={Indian Institute of Technology Bombay, Powai},
            city={Mumbai},
            country={India}}

\cortext[cor]{Corresponding author: Email: alankar.alankar@iitb.ac.in, Tel.: +91-9769415356,\\ Fax: +91-22-25726875}


\begin{abstract}
High-energy materials (HEMs) are critical for propulsion and defense domains, yet their discovery remains constrained by experimental data and restricted access to testing facilities. This work presents a novel approach toward high-energy molecules by combining Long Short-Term Memory (LSTM) networks for molecular generation and Attentive Graph Neural Networks (GNN) for property predictions. We propose a transformative embedding space construction strategy that integrates fixed SHA-256 embeddings with partially trainable representations. Unlike conventional regularization techniques, this changes the representational basis itself, reshaping the molecular input space before learning begins. Without recourse to pretraining, the generator achieves 67.5\% validity and 37.5\% novelty. The generated library exhibits a mean Tanimoto coefficient of 0.214 relative to training set signifying the ability of framework to generate a diverse chemical space. We identified 37 new super explosives higher than 9 $km. s^{-1}$ predicted detonation velocity.
\end{abstract}





\end{frontmatter}



\section{Introduction}
\noindent
The vastness of chemical space poses both a challenge and an opportunity for discovering compounds with improved properties of interest. In drug discovery, much of the challenges come from unexplored regions that could hold biologically active molecules. Many groundbreaking discoveries in chemical science have historically relied on conventional methods. Discovery of first man-made antibiotic Arsphenamine (Salvarsan) by Paul Ehrlich in 1907 was driven by screening over 600 synthetic compounds \cite{HUTCHINGS201972}. This drug proved to be crucial in treatment of syphilis, saving millions of lives. In contrast, molecular discovery has also yielded high-energy materials (HEMs), which store substantial chemical energy that can be rapidly released as heat or kinetic energy \cite{agrawal2010high}.

Material discovery of explosives and propellants has traditionally relied on resource-intensive experiments conducted within defense laboratories, leading to slow progress compared to other areas of materials research \cite{doi:10.1021/acsomega.4c01070,HUANG2021102240}. While virtual screening has long been used to prioritize candidates, advancements in machine learning (ML) and artificial intelligence (AI) now enable rapid exploration of chemical space and de novo molecular design as talked about in works of Meyers et al. \cite{MEYERS20212707}.

Molecular representation is central to data driven discovery of novel molecules \cite{doi:10.1126/science.aat2663}. Common approaches include text-based encoding such as Simplified Molecular Input Line Entry System (SMILES) and graph based representations, both of which can be readily used in generative models \cite{BAILLIF2023102566,gaudelet2021utilizing,GRISONI2023102527,MEYERS20212707,HANSER2023102545,wang2023scientific,ISERT2023102548,THOMAS2023102559}. Although SMILES are widely adopted, they are not unique representation and often result multiple valid string for same molecule. Alternate descriptors include \cite{zhang2023equivariant}, Extended connectivity fingerprint \cite{doi:10.1021/ci100050t} and molecule images \cite{zeng2022accurate}.

Generative AI has been revolutionizing the field of material and drug-discovery in recent years. In molecular design, generative models like RNN \cite{sherstinsky2020fundamentals}, VAE \cite{kingma2013auto}, GAN \cite{goodfellow2020generative}, Adversarial Autoencoder (AAE) \cite{makhzani2015adversarial} continued to be used with various modifications. LSTM--RNN have been the most frequently employed framework due to their capabilities, which reduce the likelihood of vanishing gradients. Compared to GANs, RNN-LSTM architecture based models experience mode collapse less frequently \cite{martinelli2022generative}. One of the first attempts to produce a generative model in drug discovery was by Olivercrona et al. \cite{olivecrona2017molecular} which improved efforts in molecular graph generation with RNN. The VAE-based automatic chemical design system by Gomez-Bombarelli et al. \cite{gomez2018automatic} demonstrated the use of VAEs in molecular generation. Kadurin et al. \cite{kadurin2016cornucopia} proposed the method of using a generative adversarial autoencoder (Gen-AAE) to identify fingerprints of new molecules with desired properties. Polykovskiy et al. \cite{polykovskiy2018entangled} developed conditional AAE based high-throughput generative models to discover novel molecular structures. A significant study by Yang et al. \cite{Yang31122017} showed that RNNs could innovate molecular structures with superior efficiency to that of VAEs. Similarly Van Deursen et al. \cite{van2020gen} showed that LSTM layers achieved a higher rate of valid generation—that is, the proportion of syntactically and chemically valid SMILES strings generated—compared to (Gated Recurrent Unit) GRU layers. Additionally it also showed that bi-directional GRU layers performed poorly when compared to both Uni and bi--directional LSTM. In a work on generative framework in low-data regimes, Moret et al. \cite{moret2020generative} showed that RNNs could generate valid molecules independent of explicitly provided chemical rules. From the works of Grisoni et al. \cite{doi:10.1021/acs.jcim.9b00943}, it is evident that uni--directional and bi--directional generative RNN performed significantly better than aforementioned models. Moreover it is to be noted that forward RNN outperformed bidirectional RNN in terms of valid molecule generation and had novelty much similar to bidirectional RNN.

In the works of Merk et al. \cite{https://doi.org/10.1002/minf.201700153}, RNN was trained on large dataset of molecules then fine-tuned on smaller library of molecules with desired properties. The generated dataset had 93\% valid and 90\% unique SMILES sequences with none of the generated chemical structures similar to original database.

Unlike the aforementioned approaches where a substantial amount of data and computational resources are needed for faster pretraining models and subsequently fine tuning them on selective database, our approach not only works with small data of 56.9 KB (303 rows), it requires only commodity hardware resources (Google Colab). Our model is trained within minutes and has high effectiveness comparable to large pretrained models. We focus on recent approaches in Representation Learning, particularly the application of GNNs ~\cite{wu2020comprehensive}, where molecular structures are directly used as input, eliminating the need for handcrafted feature engineering. Traditional neural architectures such as CNNs, RNNs, and GANs often rely on predefined feature representations, which can limit their ability to capture the complex relational and topological properties of molecules. GNNs address this limitation by learning representations directly from molecular graphs, offering a more flexible and chemically meaningful framework. On the contrary, Jiang et al. ~\cite{jiang2021could} reported that, on average, descriptor-based models still outperform graph-based models in predicting a diverse range of molecular properties, both in terms of predictive accuracy and computational efficiency.

Ritting et al. ~\cite{Rittig_2023} address the aforementioned challenges to accurately predict molecular properties, particularly highlighting the limitations of traditional machine learning approaches such as Quantitative Structure–Property Relationship (QSPR) models. QSPR models rely on predefined molecular descriptors that require substantial domain expertise, and the quality of prediction strongly depends on the choice of these descriptors. Inadequate or non-informative descriptors can lead to suboptimal model performance. In contrast, GNNs offer a representation learning framework that eliminates the need for manual feature engineering by directly utilizing molecular structures represented as graphs. In this formulation, atoms and bonds are represented as nodes and edges respectively, and GNNs propagate and aggregate information through a message-passing mechanism. This end-to-end learning paradigm enables GNNs to achieve outstanding performance in structure-property correlation tasks.

Waqar et al. in \cite{doi:10.1021/acsomega.2c06702}  demonstrate a robust and promising method for molecular property prediction for drug discovery. By leveraging an attention-based GNN architecture AttentiveFP \cite{doi:10.1021/acs.jmedchem.9b00959}, this method excels in representing intricate molecular structures. Traditional QSPR methods, though useful, are heavily reliant on domain specific descriptors, although they help in inference of model prediction. In contrast, attention--based GNNs, such as those explored in works \cite{doi:10.1021/acsomega.2c06702}, facilitate end-to-end learning from raw molecular graph data, reducing dependency on predefined features and offering a more nuanced molecular representation.

One of the pioneering works by Li et al. \cite{doi:10.1021/acs.jcim.2c00997} focuses on developing a generative framework for generating high energy molecules under low data constraints. The objective of the current work is to create a generative framework suitable for areas with limited data availability and low computational resources. Building on this foundation, we demonstrate a framework that uses limited data of energy materials and is able to predict novel next generation materials of similar class. Our approach employs Attention-mechanism based GNN to accurately predict the properties of generated molecules.

\label{intro}

\begin{figure}[H]
    \centering
    \begin{subfigure}[b]{0.4\textwidth}
        \centering
        \includegraphics[width=\textwidth]{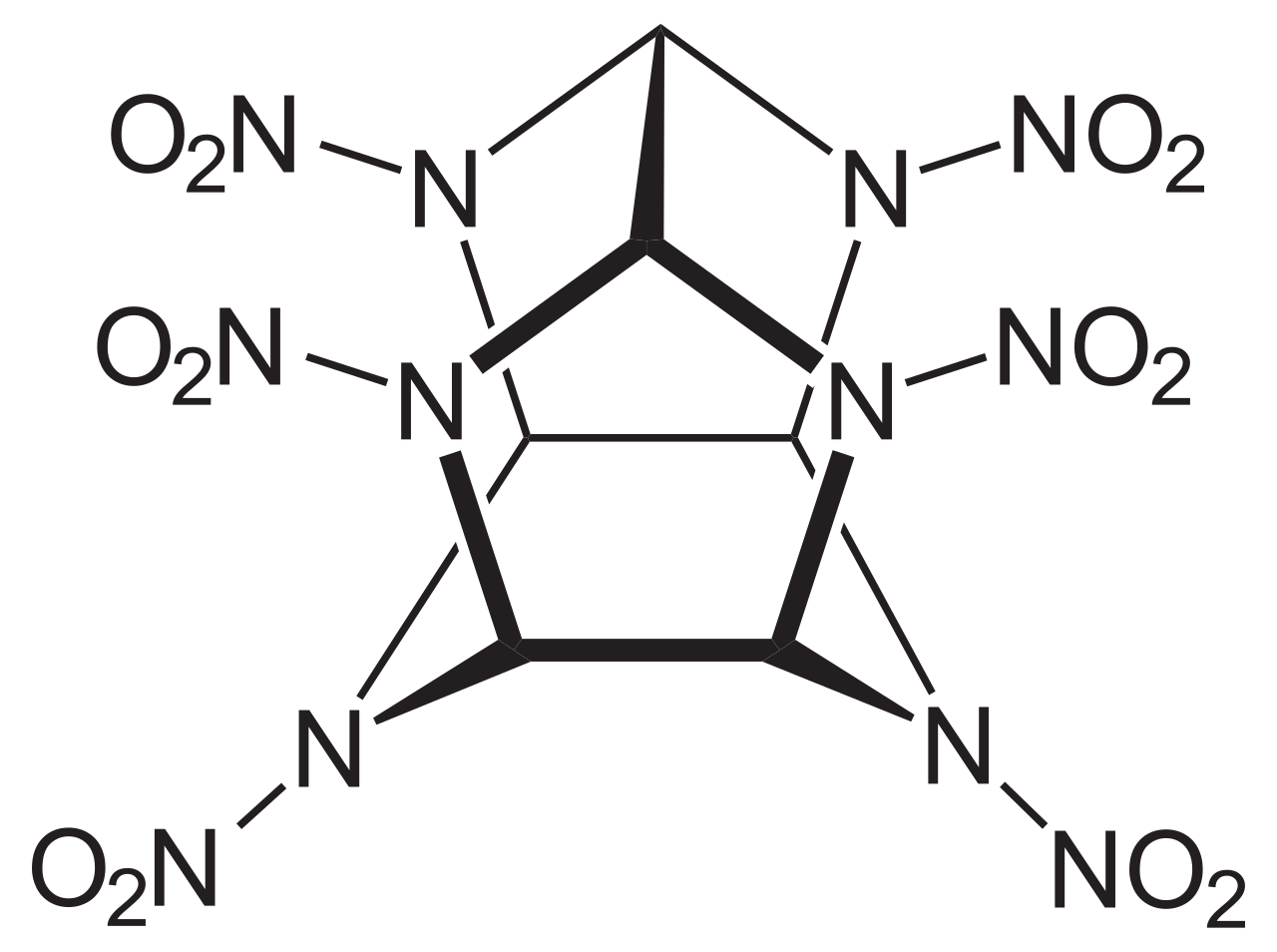}
        \caption{Visualization of molecule (2,4,6,8,10,12-Hexanitro-2,4,6,8,10,12-hexaazaisowurtzitane) 
        from original dataset.}
        \label{fig:molecule}
    \end{subfigure}
    \hfill
    \begin{subfigure}[b]{0.45\textwidth}
        \centering
        \includegraphics[width=\textwidth]{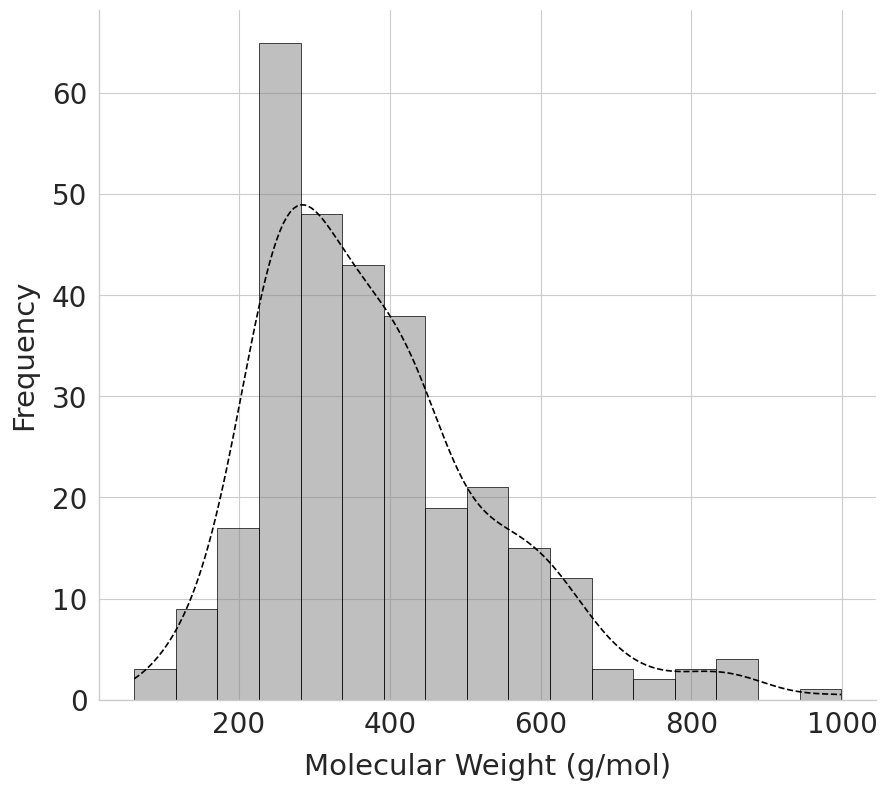}
        \caption{Histogram of weight of molecules from dataset.}
        \label{fig:histo1}
    \end{subfigure}
    
    \caption{(a) Molecule structure and (b) Molecular weight histogram.}
    \label{fig:combined}
\end{figure}

\section{Dataset}
\begin{figure}[h] 
    \centering
    \begin{subfigure}[t]{0.45\textwidth} 
        \centering
        \includegraphics[width=\textwidth]{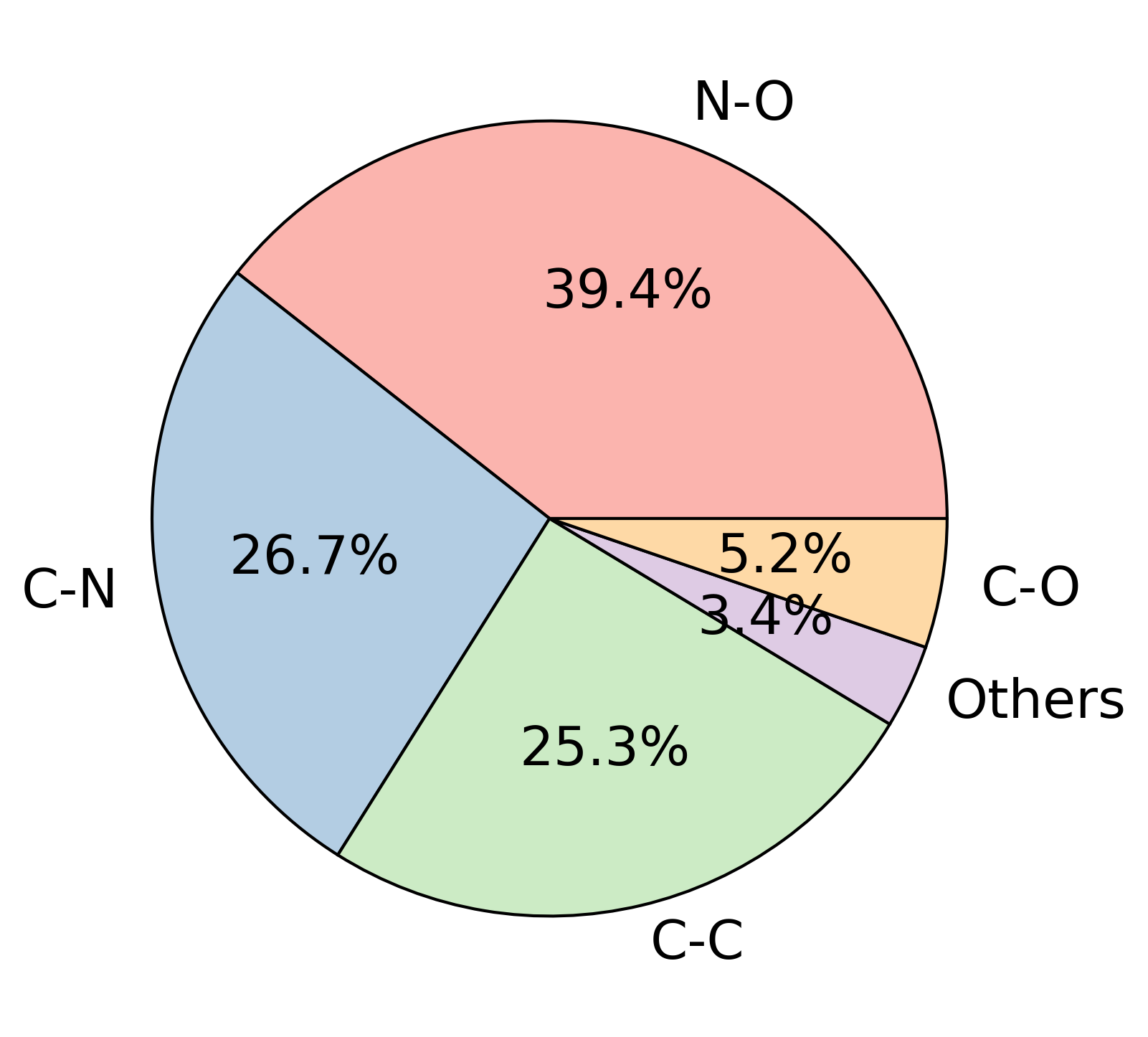} 
        \caption{}
        \label{fig:bond_distribution}
    \end{subfigure}
    \hfill 
    \begin{subfigure}[t]{0.45\textwidth} 
        \centering
        \includegraphics[width=\textwidth]{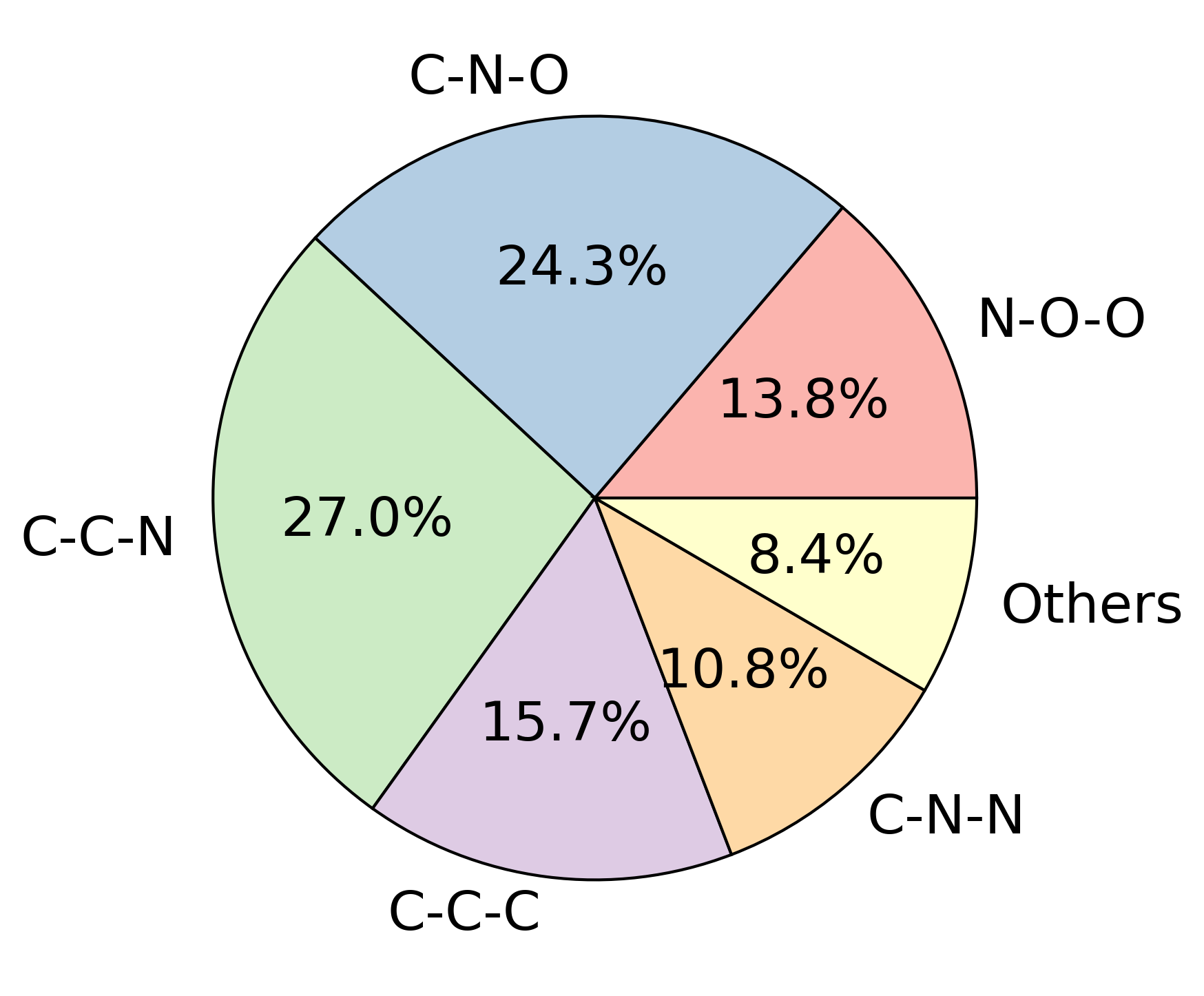} 
        \caption{}
        \label{fig:triplet_distribution}
    \end{subfigure}

    \caption{Comparison of X-X bond distribution (a) and X-X-X bond distribution (b) across the dataset.}
    \label{fig:comparison}
\end{figure}

\begin{figure}[h!]
    \centering
    \includegraphics[width=0.7\textwidth]{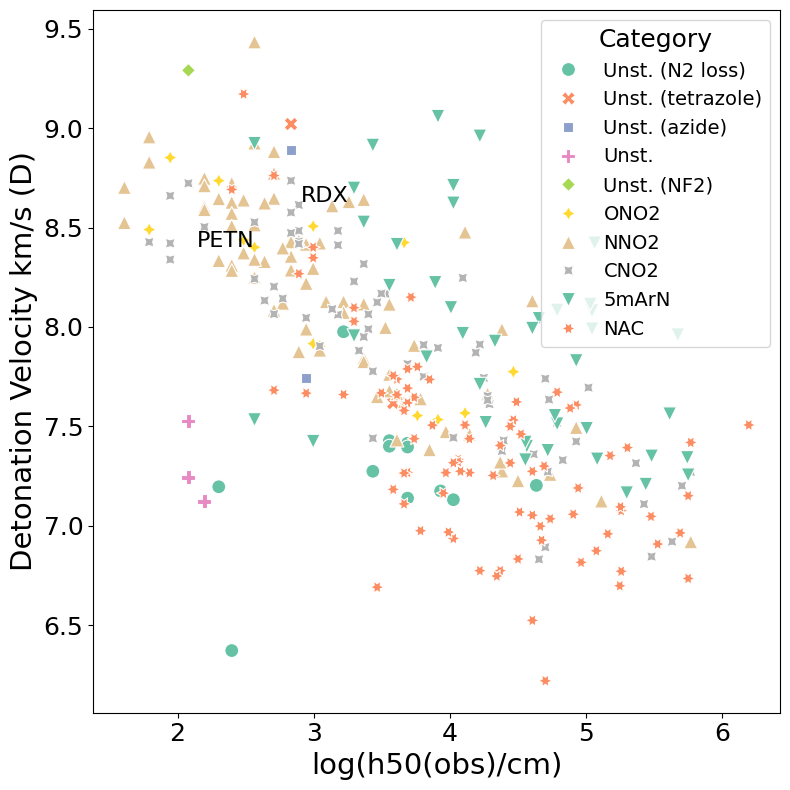}  
    \caption{The scatter plot above describes the correlation between experimental impact sensitivity and detonation velocity across different energetic material categories with \textit{log($h_{50}(obs))(cm)$} on the x-axis versus detonation velocity ($km. s^{-1}$, \textit{D}) on the y-axis. `obs' refers to observation. Distinct colors and marker shapes denote functional categories of energetic groups. RDX and PETN have been labeled for reference. The trend highlights the trade-off between higher detonation performance and reduced mechanical stability.}
    \label{fig:Detonation Velocity vs Log(h50(obs)) plot}
\end{figure}

This work utilizes the dataset of energetic compounds like the one shown in Figure \ref{fig:molecule}. The database of high energy molecules is derived from works of Mathieu et al. \cite{doi:10.1021/acs.iecr.7b02021} and Li et al.  \cite{doi:10.1021/acs.jcim.2c00997}, which contains 303 unique energetic compounds with their properties labeled. These compounds are characterized by their chemical structures and associated properties, such as detonation velocity. The dataset is relatively small, reflecting the challenge of obtaining extensive experimental data for energetic materials. As evident from the Figure \ref{fig:histo1}, the dataset of molecules used in this study has majority of molecules within 200-600 molecular weight, it is challenge to create larger molecules from the given distribution. Moreover from Figure \ref{fig:bond_distribution} and \ref{fig:triplet_distribution} it is evident that the distribution of majority of chemical bonds in high energy molecules is quite different from those in any chemical space. Motivated by this we avoid pretraining which harms the weights creating bias in generative model leading to generating molecules of undesired properties. Our main focus is to use the \textit{lack of data} to our advantage such that the unconditional learning in restrictive chemical space as visible in Figure \ref{fig:Detonation Velocity vs Log(h50(obs)) plot} enables it to learn to generate from those same chemical spaces.

\subsection{Molecular Descriptors and Properties in the Dataset}

The dataset contains essential molecular descriptors and physical properties of high-energy materials, providing insights into their chemical structure, behavior, and performance. Each molecule is represented by its SMILES string, which offers a convenient, text-based encoding of the molecular structure. The key features in our dataset are described in Table \ref{table:metrics}.

\begin{table}[h!]
\caption{Description of molecular properties and metrics.}
\label{table:metrics}
\centering
 \small
\begin{tabular}{p{2.5cm} p{12cm}}
\hline
\textbf{Column} & \textbf{Description} \\ \hline
\textbf{SMILES} & SMILES string is a linear notation that encodes the structure of the molecule using short ASCII text, enabling the efficient representation of complex chemical structures. \\ \hline

\textbf{Category} & Categorizes the molecules, likely based on their chemical functionality or energetic properties (e.g., explosives, propellants). \\ \hline
\textbf{OB(CO$_2$)} & Oxygen balance (OB) with respect to CO$_2$. It provides a measure of the stoichiometric ratio of oxygen required to fully oxidize the molecule’s carbon and hydrogen atoms, important for evaluating combustion or detonation efficiency. \\ \hline
\textbf{r$_0$} & The initial density of the molecule, typically measured in grams per cubic centimeter (g/cm$^3$). This directly influences its performance, especially in detonation velocity and pressure. \\ \hline
\textbf{HGAS} & The heat of gasification, representing the energy required to convert the molecule from a liquid or solid state into a gaseous state, influencing behavior under high-temperature conditions. \\ \hline
\textbf{HSUB} & The heat of sublimation, quantifying the energy required for a substance to transition directly from a solid to a gaseous state. It is essential for understanding the thermal stability of high-energy materials. \\ \hline
\textbf{Q} & The heat of reaction (or energy release) during combustion or detonation. This determines the power and efficiency of an explosive material. \\ \hline
\textbf{D} & Detonation velocity, typically measured in meters per second ($km. s^{-1}$), representing the speed at which a detonation wave propagates through the material, indicating explosive performance. \\ \hline
\textbf{P} & Detonation pressure, commonly measured in gigapascals (GPa), which indicates the peak pressure generated during detonation, correlating with destructive potential. \\ \hline
\textbf{EG} & Gurney energy of a molecule measures the kinetic energy imparted to fragments or gases upon detonation, critical for optimizing the performance of explosives. \\ \hline
\textbf{h$_{50}$(obs)} & The impact sensitivity of the material, measured as the height at which the material has a 50\% probability of detonation upon impact. A lower $h_{50}$ value indicates higher sensitivity to mechanical shock, making this an important safety parameter. obs refers to observed quantity. \\ \hline
\end{tabular}

\end{table}

In summary, these descriptors provide a comprehensive understanding of the molecular structure and physical properties of high-energy materials. They are crucial for predicting the behavior of these compounds in various applications, including their detonation characteristics, safety considerations, and overall energetic performance.

\section{Methodology}

\subsection{Data augmentation}
To enhance the diversity of molecular representation we have employed data augmentation by randomizing SMILES sequences by enumeration as described in works of \cite{bjerrum2017smilesenumerationdataaugmentation}. We have conducted controlled experiments on $1\times$, $3\times$, $5\times$ augmentation and compared their performances.

\subsection{Generation model}
\begin{figure}[h!]
    \centering
    \includegraphics[width=\textwidth]{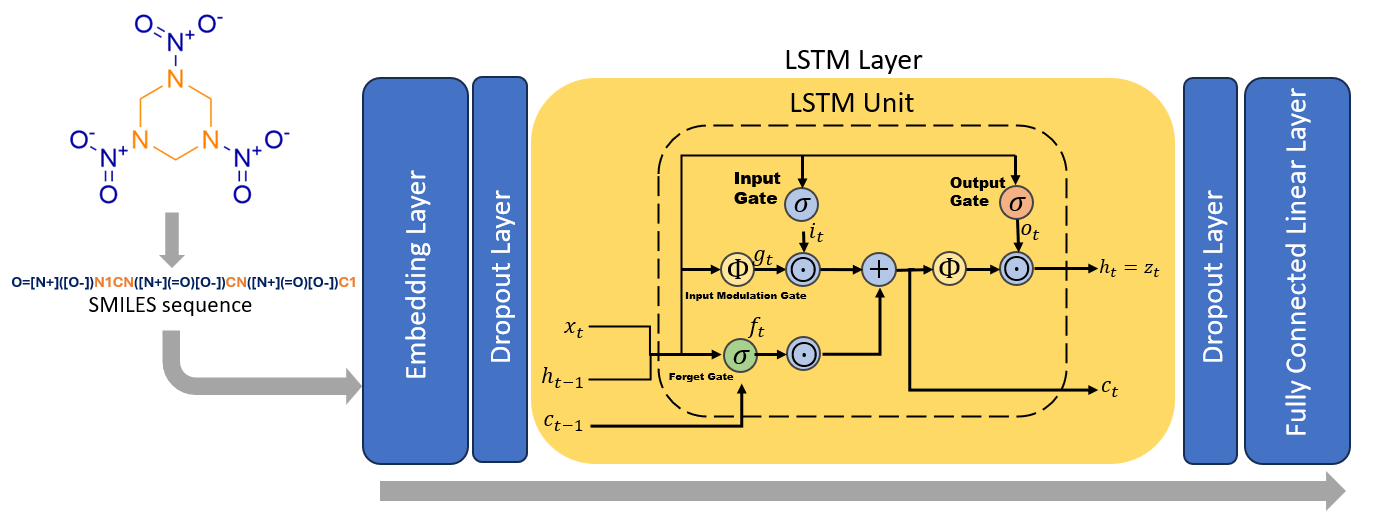}
    \caption{Schematic overview of the generative model architecture used for high-energy molecule generation. The molecular structure is first represented using the SMILES sequence which is then tokenized and passed through an embedding layer. A dropout layer introduces regularization to mitigate overfitting in the low-data regime. The core LSTM architecture processes the sequential data, where each LSTM unit updates hidden states through the interaction of input, forget, and output gates as per standard LSTM dynamics. The processed latent representation is further passed through another dropout layer and a fully connected linear layer to decode the next character in the sequence. This lightweight model architecture facilitates rapid generation of valid, novel, and diverse high-energy molecules using minimal computational resources.%
    }
    \label{fig:Model architecture}
\end{figure}

LSTM-based methods have shown strong performance in molecule generation with high validity and novelty \cite{santana2020novo,howard2018universal}. However, their ability to capture the desired chemical space remains uncertain. The current study finds that ChEMBL-pretrained approaches \cite{doi:10.1021/acs.jcim.2c00997,li2020inductive,singh2022transfer}, despite heavy computational cost, often generate high-energy molecules that resemble those produced by a simple LSTM. This suggests that competitive generation quality can be achieved without pretraining, though at the expense of novelty, which requires multiple iterations to accumulate a substantial database of molecules. Motivated by this, we explore a new generation model designed to use far fewer resources. In our work we are engineering the embedding layer of the architecture for better learnability. Except the embedding layer, the model architecture stays the same as described in Figure \ref{fig:Model architecture}.  

\subsubsection{Training process}

The model was trained using the Adam optimizer, with systematic variation of hyperparameter to evaluate robustness across different configurations. The learning rate (\( \text{lr} \)) was explored over four values: \(1 \times 10^{-4}\), \(1 \times 10^{-3}\), \(5 \times 10^{-3}\), and \(1 \times 10^{-2}\). Dropout rates of 0.1, 0.3, 0.5, and 0.7 were examined. Batch sizes of 16, 32, and 64 were tested. Further evaluation with training dimensions of 10, 30, 50, 80, and 100 for model 2 and model 3 was done.  

Training was conducted for 300 epochs, as preliminary trials with up to 900 epochs led to significant overfitting, reflected in reduced novelty of generated sequences. Cross-entropy loss was used as the objective function, consistent with the multi-class classification nature of next-token prediction in SMILES strings. A custom collate function was implemented to dynamically manage padding within batches of variable-length sequences.  

During optimization, the network was trained to predict the next character given the preceding sequence. Model performance was assessed at the end of each epoch using the coefficient of determination (\(R^2\)), mean absolute error (MAE), and L2 loss, computed for both training and validation sets to monitor convergence and detect overfitting across hyperparameter regimes.

\subsubsection{Model Architectures}
The architectures explored for SMILES generation can be seen below :
\begin{enumerate}
    \item \textit{Model 1: Vanilla LSTM with two layers}\\
    This model uses two unidirectional LSTM layers with and without dropout and a standard learnable embedding.

    \item \textit{Model 2: Partially trainable-embedding LSTM with two layers}\\
    This model uses two unidirectional LSTM layers with and without dropout and a partially trainable embedding created by concatenating a non-trainable embedding with a trainable one.

    \item \textit{Model 3: Hash-embedding LSTM with two layers}\\
    This model uses two unidirectional LSTM layers with and without dropout and a partially trainable embedding created by concatenating a non-trainable Secure Hashing Algorithm (SHA-256)~\cite{FIPS180-2}-based embedding with a trainable one.

    \item \textit{Model 4: Universal Language Model Fine-tuning (ULMFiT)}\\
    Baseline model used in our study for comparison.
\end{enumerate}

We are using the four different model architectures to systematically explore the effect of embedding strategies and transfer learning on SMILES generation.

\subsubsection{Partially‐trainable SHA‐256 Token embeddings}  
To address overfitting and data scarcity in SMILES generation, we employ a hybrid embedding matrix \(E \in \mathbb{R}^{V\times d}\), where only a subspace \(E^{(t)}\in\mathbb{R}^{V\times d_t}\) is trainable, and the remaining \(E^{(f)}\in\mathbb{R}^{V\times d_f}\) (\(d_f=d-d_t\)) is deterministically initialized via SHA‑256 hashing \cite{FIPS180-2} as shown in Algorithm \ref{alg:sha256_embedding}. For each token index \(i\), we compute
\[
h_i = \mathrm{SHA256}(\texttt{token(i)}),\quad 
b_i = \mathrm{bytes}(h_i),\quad 
v_i = \frac{b_i - 128}{128}\in[-1,1]^{32},
\]
then tile or truncate \(v_i\) to length \(d_f\), forming the fixed embedding \(E^{(f)}[i]\), and set \(E[i]=[\,E^{(t)}[i]\parallel E^{(f)}[i]\,]\).

Since \(E^{(f)}\) is not back-propagated, it helps reduce usage of GPU memory and runtime \cite{pmlr-v37-chenc15}. Its deterministic nature ensures reproducibility. Moreover, the fixed component acts as a new learning representation, similar to the NLP “hashing trick” approach \cite{NIPS2017_f0f6ba4b,bojanowski-etal-2017-enriching,NEURIPS2018_7e837225}. SHA‑256 offers high-entropy, collision-resistant, reproducible encoding, introducing non-learned entropy without redundant noise. Character-level SMILES embeddings often suffer token sparsity. The SHA‑256 method ensures diverse and deterministic encoding, preventing early syntactic memorization and enhancing chemical-space exploration \cite{joulin2016bagtricksefficienttext}. This is particularly crucial when pretrained models (e.g., ULMFiT or ChEMBL-finetuned) fail to generalize effectively to high-energy molecule domains. This partially-trainable embedding technique is not only shown to be empirically improving novelty and diversity in SMILES generation under limited-data conditions but also theoretically justified in supporting information.

\begin{algorithm}[ht]
\caption{Generate SHA-256-Based Fixed Embeddings. See Appendix \ref{appendixA} for detail.}
\label{alg:sha256_embedding}
\begin{algorithmic}[1]
\Require Vocabulary size $V$, embedding dimension $d$, trainable dimension $d_t$ where $d_t \leq d$
\Ensure Fixed embedding matrix $W_f \in \mathbb{R}^{V \times (d - d_t)}$
\State $d_f \gets d - d_t$
\State Initialize $W_f \gets \mathbf{0}^{V \times d_f}$
\For{$i = 0$ to $V-1$}
    \State $s \gets \text{UTF-8 encoding of string ``token(i)}\text{''}$
    \State $h \gets \text{SHA256}(s)$ \Comment{$h \in \{0,1\}^{256}$}
    \State $b \gets \text{Byte array from } h$ \Comment{$b \in \{0, \dots, 255\}^{32}$}
    \State $v \gets \left( \text{float32}(b) - 128.0 \right) / 128.0$ \Comment{Normalize to $[-1, 1]$}
    \State $v' \gets \text{Repeat or trim } v \text{ to size } d_f$
    \State $W_f[i] \gets v'$
\EndFor
\end{algorithmic}
\end{algorithm}

\subsubsection{Evaluation Metrics}

To evaluate the performance of the SMILES generator and the associated predictive models, we employed a combination of regression metrics (for property prediction) and generative quality metrics (for SMILES outputs). The metrics are described as follows. 

\textbf{\( R^2 \) Score:}

\begin{equation}
R^2 = 1 - \frac{\sum_{i=1}^{N} (y_i - \hat{y}_i)^2}{\sum_{i=1}^{N} (y_i - \bar{y})^2}
\end{equation}

Here, \( y_i \) denotes the ground-truth property value of the \( i \)-th molecule, \( \hat{y}_i \) is the corresponding predicted value, and \( \bar{y} = \frac{1}{N} \sum_{i=1}^{N} y_i \) is the mean of observed property values. The \( R^2 \) score measures the proportion of variance in molecular property predictions explained by the model. Higher \( R^2 \) values indicate better performance.
\newline

\textbf{Validity:}

Let \( \mathcal{G} = \{ s_1, s_2, \dots, s_n \} \) denote the set of generated SMILES strings. A molecule \( s_i \in \mathcal{G} \) is considered valid if it can be parsed into a chemically valid structure using RDKit \cite{landrum2013rdkit}. Validity is computed as

\begin{equation}
\text{Validity} = \frac{1}{n} \sum_{i=1}^{n} \mathbb{I}\big[\texttt{is\_valid}(s_i)\big]
\end{equation}

where \( \mathbb{I}[\cdot] \) is the indicator function that returns 1 if \( s_i \) is valid, and 0 otherwise.

\textbf{Novelty:}

Let \( \mathcal{D}_{\text{train}} \) be the training set molecules and \( \mathcal{G}_{\text{valid}} \subseteq \mathcal{G} \) the valid generated molecules. A molecule \( s_i \) is novel if it does not occur in \( \mathcal{D}_{\text{train}} \). Novelty is defined as

\begin{equation}
\text{Novelty} = \frac{1}{|\mathcal{G}|} 
\sum_{s_i \in \mathcal{G}_{\text{valid}}} 
\mathbb{I}\big[s_i \notin \mathcal{D}_{\text{train}}\big]
\end{equation}

\textbf{Uniqueness:}

Uniqueness measures the proportion of non-duplicate molecules among the valid set. Let \( \text{set}(\mathcal{G}_{\text{valid}}) \) denote the set of canonicalized unique molecules. Then

\begin{equation}
\text{Uniqueness} = \frac{|\text{set}(\mathcal{G}_{\text{valid}})|}{|\mathcal{G}_{\text{valid}}|}
\end{equation}

\subsection{Prediction Model}
\begin{figure}[h!]
    \centering
    \begin{subfigure}[b]{0.9\textwidth}
        \centering
        \includegraphics[width=\textwidth]{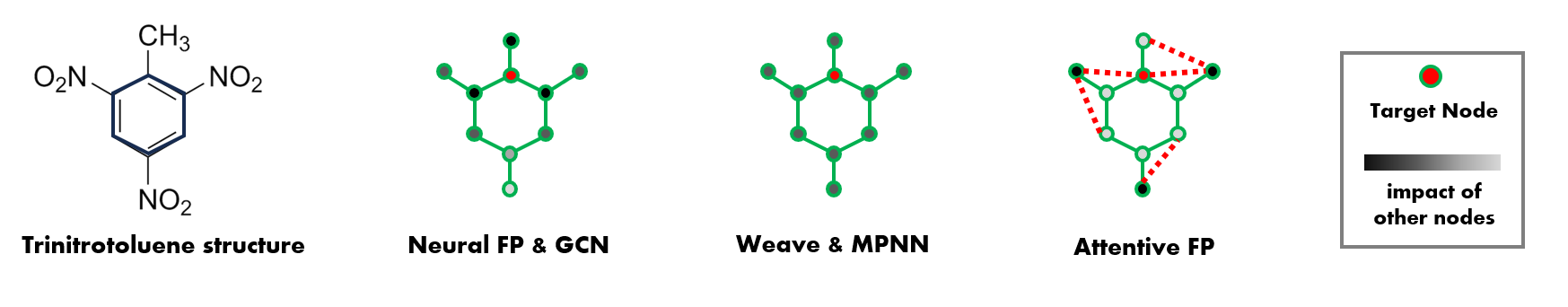}
        \caption{}
        \label{fig:Graph based heuristic1}
    \end{subfigure}
    \hfill
    \begin{subfigure}[b]{0.9\textwidth}
        \centering
        \includegraphics[width=\textwidth]{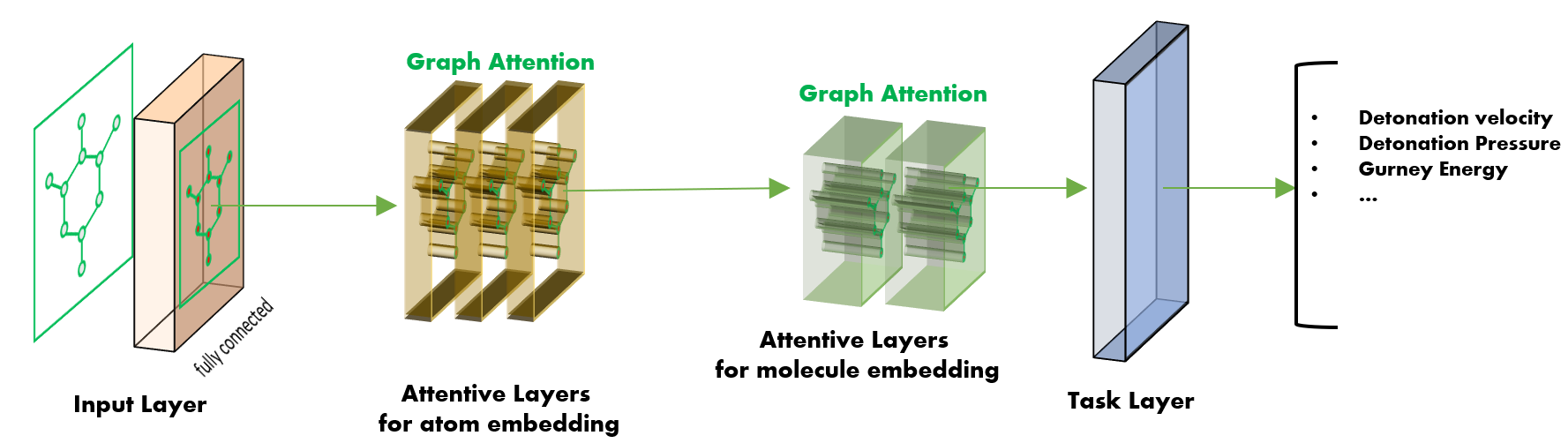}
        \caption{}
        \label{fig:Graph based heuristic2}
    \end{subfigure}
    \caption{(a) Graph-based heuristic for feature extraction, and (b) schematic of a graphical neural network with attention.}
    \label{fig:graph_heuristic_gnn}
\end{figure}

We use GNN for predicting multiple molecular properties of generated molecules. We have leveraged AttentiveFP \cite{doi:10.1021/acs.jmedchem.9b00959} architecture for capturing both molecular topological changes and feature relationships , a brief schematic of feature mapping is shown in Figures \ref{fig:Graph based heuristic1} and \ref{fig:Graph based heuristic2}.

\begin{figure}[h!]
    \centering
    \includegraphics[width=1\textwidth]{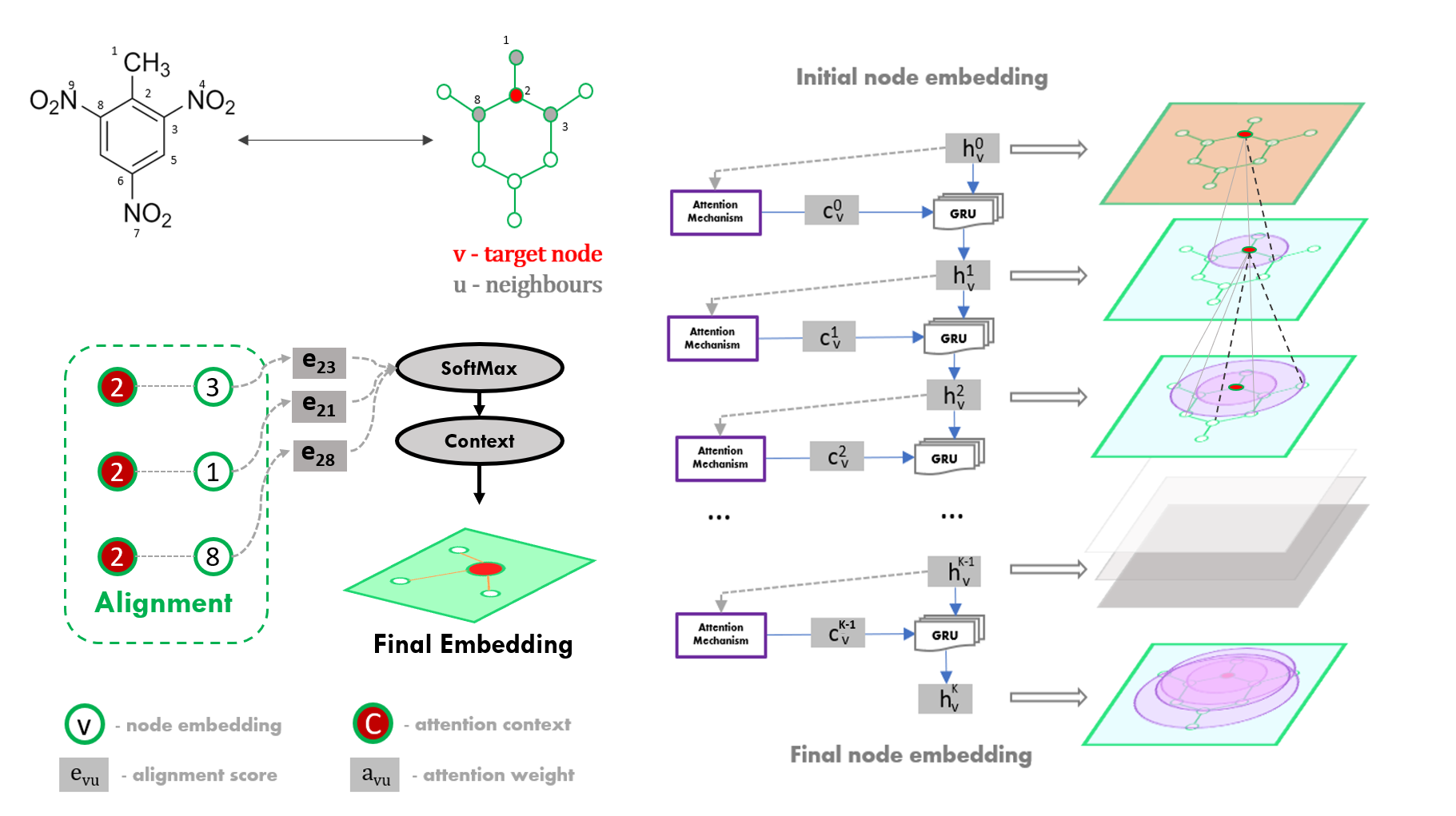}  
    \caption{Overview of Attention-based message passing in molecular graphs. For each target atom $v$ (red), neighbor embeddings $u$ are aggregated using attention scores and weights to form a context vector. This context is iteratively updated through GRUs across layers to produce refined node embeddings capturing the atom’s chemical environment.}
    \label{fig:Graph Attention Mechanism1}
\end{figure}

\subsubsection{Dataset preparation for prediction model}
Each molecule's structure is represented as a graph, where nodes correspond to atoms, and edges represent chemical bonds. The dataset includes multiple target properties for each molecule, which are vital for multi--target prediction. These properties, namely \texttt{OB(CO2)}, \texttt{r0}, \texttt{HGAS}, \texttt{HSUB}, \texttt{Q}, \texttt{D}, \texttt{P}, \texttt{EG}, and \texttt{h50(obs)}, are used as output targets in the model, and are each associated with a corresponding value for every molecule in the dataset. To keep the low data less sensitive, we have used StandardScaler for normalization of target values.

\subsubsection{Graph construction and multi-target output}
The molecular graphs are constructed using a RDKit function \cite{landrum2013rdkit}, which takes a SMILES string as input and converts it into a graph representation. The multi-target nature of the prediction task is handled by creating a tensor for each molecule that contains the target properties. Specifically, the output tensor for each molecule consists of a 9-dimensional vector representing the values for \texttt{OB(CO2)}, \texttt{r0}, \texttt{HGAS}, \texttt{HSUB}, \texttt{Q}, \texttt{D}, \texttt{P}, \texttt{EG}, and \texttt{h50(obs)}. These target values are used to train the model for multi-target regression.

\subsubsection{Prediction Model architecture}
We utilize the AttentiveFP model, a state-of-the-art GNN architecture designed for molecular property prediction. The model consists of several layers of graph convolutions, followed by an attention mechanism that enables it to focus on the most informative parts of the molecular graph as evident in Figure \ref{fig:Graph Attention Mechanism1}. The attention mechanism assigns a weight to each node, allowing the model to prioritize important atoms and bonds when making predictions. The model architecture is set to have 9 output channels, corresponding to the 9 target properties. The hidden layers are composed of 118 hidden channels, and the model employs a 3-layer structure with a 3-step temporal attention mechanism. This setup enables the model to efficiently capture both local and global structural features in molecular graphs (See Figure \ref{fig:Graph Attention Mechanism1}). 

\subsubsection{Training and evaluation}

For the regression-based property prediction using the AttentiveFP model, we use the mean squared error (MSE) loss as given by
\begin{equation}
\mathcal{L}_{\text{MSE}} = \frac{1}{N} \sum_{i=1}^{N} (y_i - \hat{y}_i)^2.
\end{equation}
The model is trained using the Adam optimizer \cite{kingma2017adammethodstochasticoptimization} with a learning rate of $10^{-4}$ and weight decay of $10^{-3}$. During training, we use mean squared error (MSE) loss to minimize the difference between the predicted and actual target values for each molecule. The model is trained over multiple epochs, and the root mean squared error (RMSE) computed at each epoch to evaluate the performance of the model on both the training and test datasets. 
\subsubsection{Attention mechanism in GNNs}
The attention mechanism in GNNs is designed to improve the model's ability to focus on important parts of the molecular graph. In the case of the AttentiveFP model, the attention mechanism assigns a weight to each node in the graph based on its relevance to the task. This process allows the model to dynamically adjust the importance of different atoms in the molecule, which is particularly useful when predicting complex properties that may be influenced by specific substructures. The attention weights are learned during training and are used to aggregate node features in a way that prioritizes the most informative atoms. By incorporating attention, the model can effectively learn from diverse molecular structures and provide more accurate predictions for each target property.

The model was implemented using PyTorch \cite{paszke2019pytorch}, and trained on a GPU on Google Colab to accelerate computation. The dataset and model were managed using standard data manipulation and processing libraries such as Pandas \cite{mckinney2011pandas} and NumPy \cite{harris2020array}. RDKit \cite{landrum2013rdkit}, a cheminformatics toolkit, was used for SMILES validation, structure generation, and descriptor calculations.

\section{Results}

The performance of the various generative models are analyzed by a series of simulations under the comparable experimental conditions. After running the generation Model1 and Model4 with the chosen hyperparameters. We infer that Model4, that was pretrained on the ChEMBL database and subsequently fine-tuned on our dataset, achieved a validity of up to $95\%$ and a novelty of $94\%$. In contrast, Model1 exhibited limited capability, securing a validity of only around 23\%, primarily due to the less data that often leads to overfitting. Model1 employs a lightweight RNN architecture for SMILES generation, requiring minimal computational resources, whereas Model4 utilizes a state-of-the-art generative framework that leverages significantly higher computational power to achieve superior performance. Detailed evidence and comparative analyses are provided in the supporting information.

Building upon these observations, we extended our investigation to intermediate architectures, namely Model2 and Model3, in order to explore the trade-off between model complexity, data efficiency, and molecular diversity. These models were designed to balance computational cost with generative capability, enabling a broader assessment of how architectural and hyperparameter variations influence the quality and novelty of generated molecules.

\begin{figure}[h!]
    \centering
    \begin{subfigure}[b]{0.45\textwidth}
        \includegraphics[width=\textwidth]{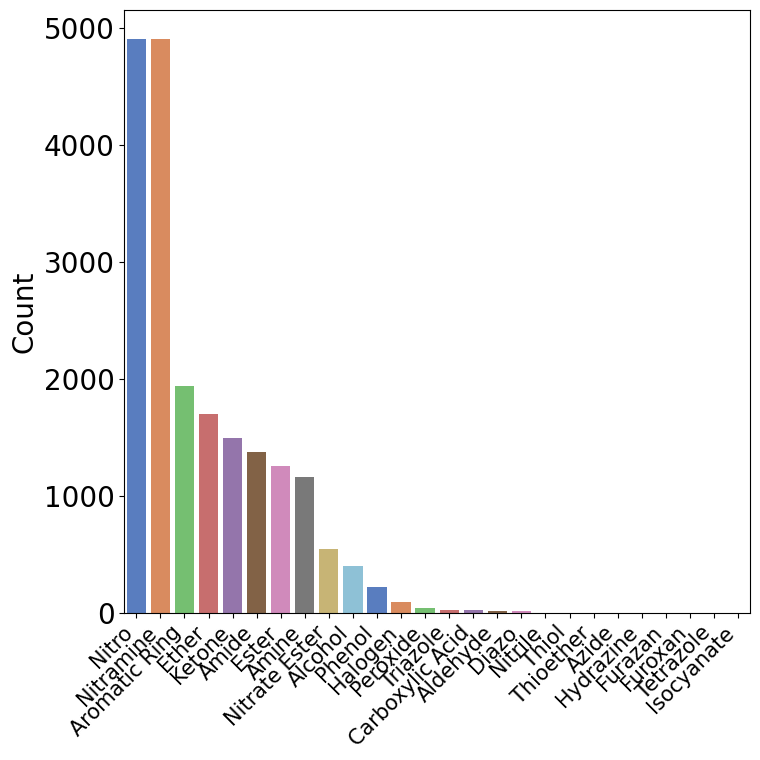}
        \caption{}
        \label{fig:functional_group_bar}
    \end{subfigure}
    \hfill
    \begin{subfigure}[b]{0.45\textwidth}
        \includegraphics[width=\textwidth]{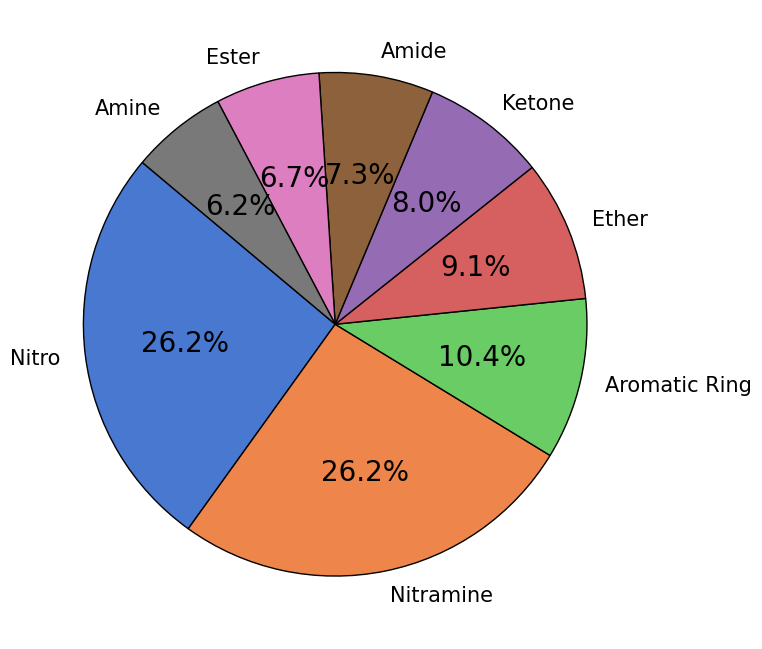}
        \caption{}
        \label{fig:functional_group_pie}
    \end{subfigure}
    \caption{Overview of functional group distribution in the molecular dataset. (a) Distribution of all functional groups (b) distribution of top 8 most frequent groups.}
    \label{fig:functional_group_combined}
\end{figure}

Through a series of controlled experiments on Model 2 and Model 3 with different hyperparameter we are able to curate database comprising total of 4964 novel molecules represented by SMILES strings. For Model 2 we are able to generate 4502 molecules out of which 2562 are novel. Similarly from Model 3 we are able to generate 4861 out of which 2762 are novel. Functional group analysis shown in Figure \ref{fig:functional_group_bar} reveals that nitro and nitramine moieties are the most abundant, each occurring in 4910 molecules, followed by aromatic rings, ether, ketones, amides and esters. These functionalities are commonly associated with energetic performance, stability and oxygen balance in explosives and propellants. A more focused analysis of the top eight functional groups shown in Figure \ref{fig:functional_group_pie} confirms that nitro and nitramine groups together constitute more than 50\% of the total functional group occurrences in the dataset.

\begin{figure}[h!]
    \centering
    \begin{subfigure}[b]{0.45\textwidth}
        \includegraphics[width=\textwidth]{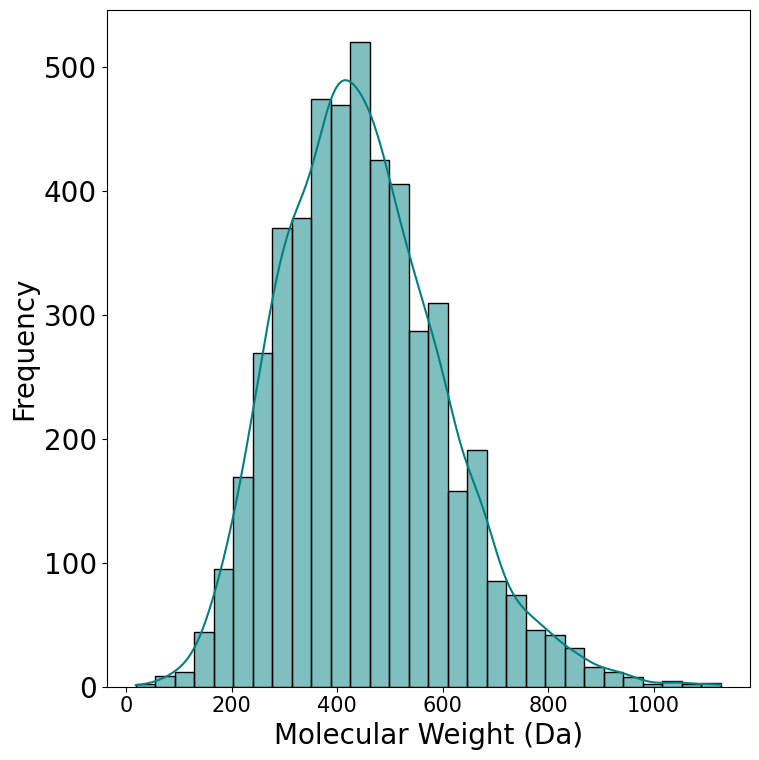}
        \caption{}
        \label{fig:molecular_weight22}
    \end{subfigure}
    \hfill
    \begin{subfigure}[b]{0.45\textwidth}
        \includegraphics[width=\textwidth]{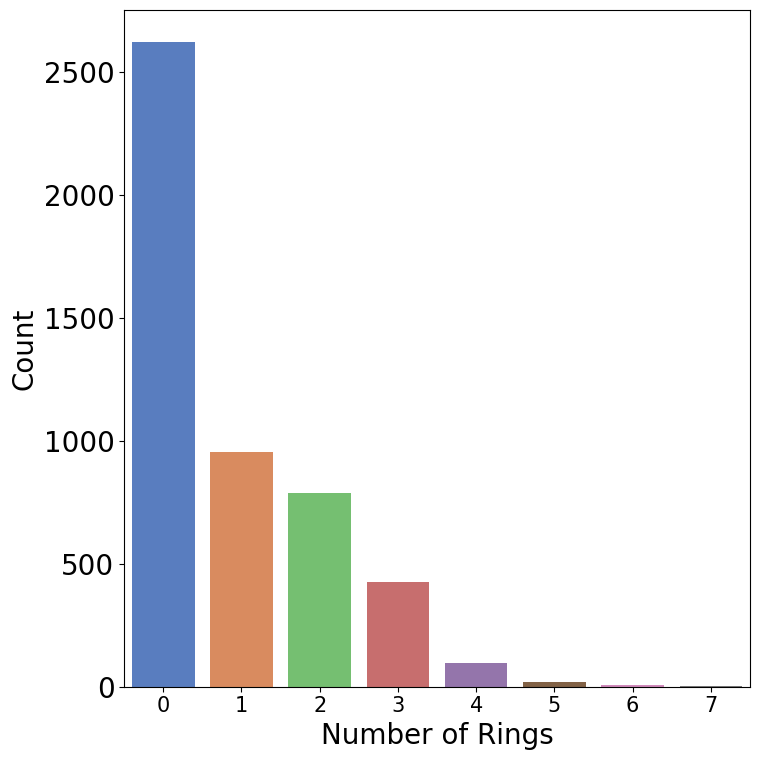}
        \caption{}
        \label{fig:ring_count}
    \end{subfigure}
    \caption{(a) Distribution of molecular weights and (b) ring systems of all generated molecular data.}
    \label{fig:molecular_and_ring}
\end{figure}

Molecular weight distribution in Figure \ref{fig:molecular_weight22} spans a range of 50 to over 1,000 Da, with majority between 400 and 500 Da. This distribution aligns with the typical size range of small-to-medium energetic compounds, balancing performance with ease of synthesis. The dataset also includes structural diversity in terms of ring systems (see Figure \ref{fig:ring_count}), where over half of the molecules are acylic and a significant portion contains one or two rings. This chemical-structural variation includes both aliphatic and aromatic frameworks, supporting a broad spectrum of potential applications, from high explosives to propellant additives.

To investigate the influence of embedding strategy on generative performance, we compared two variants of our SMILES-generating LSTM model. One variant utilized random fixed embeddings (Model 2) and other incorporated SHA-based hash embeddings (Model 3) as part of a partially trainable embedding layer. The fixed portion of the hash embedding encodes each token deterministically using SHA-256, providing a consistent and structured representation across training runs.

\begin{figure}[h!]
    \centering
    \includegraphics[width=0.5\textwidth]{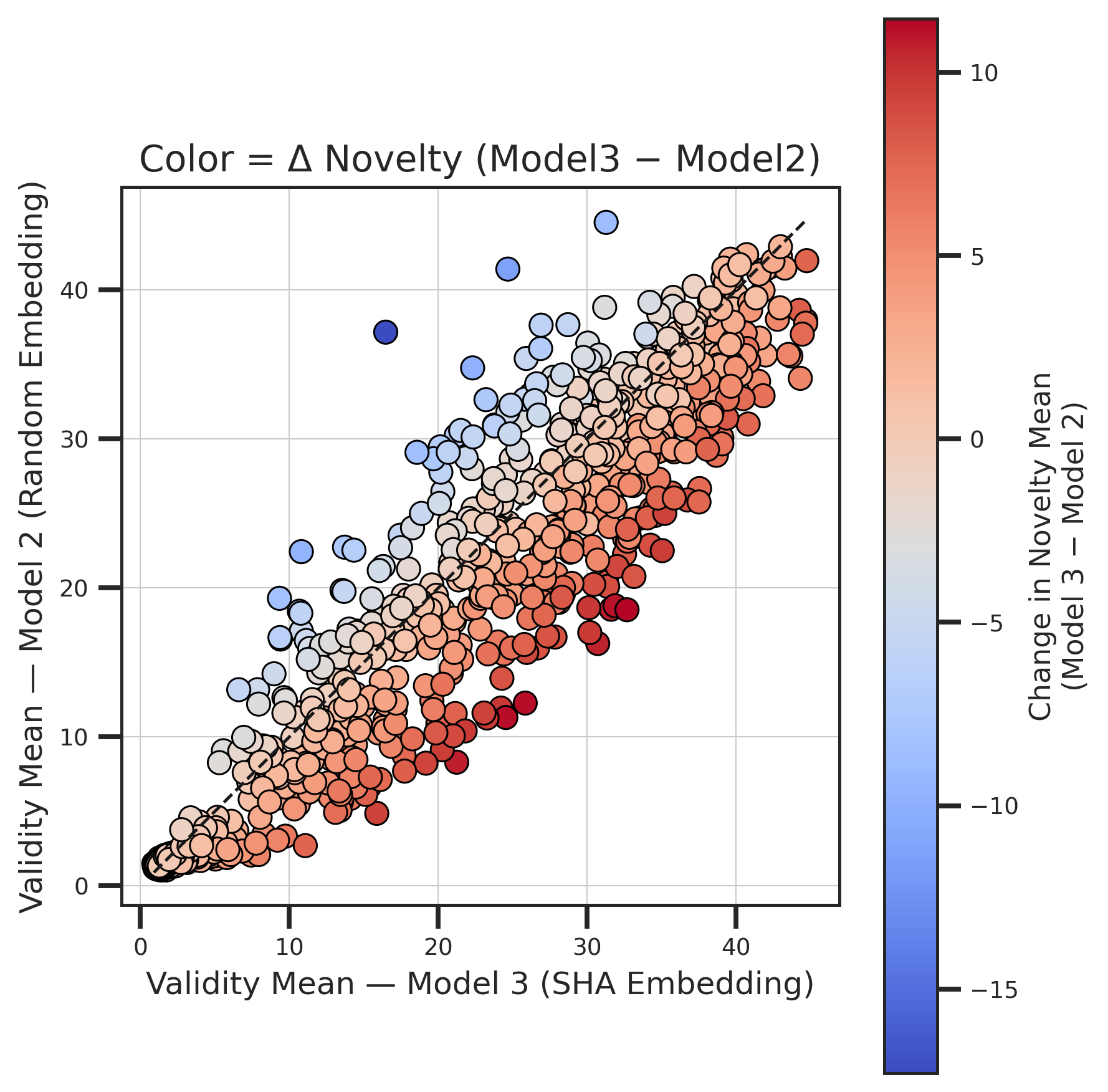}  
    \caption{Comparison of model performance across different hyperparameter configurations using two embedding strategies: random fixed embedding (Model 2) and SHA-256 hash-based fixed embeddings (Model 3). The X-axis shows the mean validity score for Model 3, and the Y-axis shows the mean validity score for Model 2. Each point is representing a different hyperparameter configuration, the coloring refers to the change in novelty shift (Model 3-Model 2). Red points indicate configurations where the hash-based embedding improves novelty, while blue indicates a drop. The diagonal line denotes parity in validity.}
    \label{Plot Comparision for Model Performance}
\end{figure}

From Figure \ref{Plot Comparision for Model Performance} we get a comparative analysis of both models across a diverse hyperparameter sweep. We observe that a significant portion of the points lie below the diagonal, indicating that Model 3 consistently outperforms Model 2 in terms of validity. This suggests that the inclusion of a deterministic, structured fixed embedding (via SHA-256) provides a more stable and generalizable inductive bias than randomly initialized fixed vectors. Notably, many configurations also exhibit a positive shift in novelty, shown by the prevalence of red-hued points.

However, this benefit is not consistent across all settings. In cases with high dropout or very small trainable embedding dimensions, Model 3 performed poorly, likely because it lacked sufficient adaptability to offset the fixed nature of the hash embeddings. Overall, our findings highlight that structural fixed embeddings- when balanced with sufficient trainable capacity ,can improve generalization and enhance the diversity of generated outputs. These results imply that SHA-based embeddings can aid the model in exploring chemically novel yet syntactically valid regions of SMILES space.

To predict physicochemical properties, we employed AttentiveFP, a graph neural network architecture that integrates attention mechanisms into the molecular fingerprinting process. This model was bench marked against several established approaches, including Hessian-regularized multilayer perceptrons (Hessian MLP), fingerprint-based Random Forest models, and two canonical graph-based models, Graph Convolution Networks (GCN) and Graph Attention Networks (GAT).

AttentiveFP demonstrated consistently superior performance across all evaluated molecular properties, as summarized in Table~\ref{tab:performance_comparison}. For example, it achieved $R^2$ values of 0.9776 for OB(CO\textsubscript{2}), 0.9046 for HGAS, and 0.6423 for h50(obs), outperforming all other models by significant margins. In contrast, baseline GNNs like GCN and GAT exhibited unstable or even negative predictive correlations, likely due to their limited expressivity and inability to capture long-range dependencies without attention mechanisms.

These results were achieved under computational constraints, including restricted GPU memory and limited batch sizes. Despite this, AttentiveFP displayed efficient convergence and robust generalization, requiring fewer epochs and minimal hyperparameter tuning.

\begin{table}[htbp]
\centering
\caption{Performance comparison ($R^2$ scores) of predictive models across molecular properties. Best performance in each row is shown in bold fonts.}
\label{tab:performance_comparison}
\resizebox{\textwidth}{!}{%
\begin{tabular}{lccccccccc}
\toprule
\textbf{Model} & OB(CO\textsubscript{2}) & r\textsubscript{0} & HGAS & HSUB & Q & D & P & EG & h50(obs) \\
\midrule
AttentiveFP & \textbf{0.9776} & \textbf{0.9800} & \textbf{0.9046} & \textbf{0.9845} & \textbf{0.6476} & \textbf{0.9174} & 0.9198 & \textbf{0.7853} & \textbf{0.6423} \\
HessianMLP            & 0.9706 & 0.9698 & 0.7347 & 0.5825 & 0.4450 & 0.9133 & \textbf{0.9249} & 0.7823 & 0.2563 \\
Fingerprint+RF        & 0.6973 & -9.4762 & 0.7264 & 0.5019 & -1.1966 & -2.1177 & 0.3313 & 0.3823 & 0.0391 \\
GCN                   & -0.0568 & -0.3155 & -0.0210 & 0.0803 & -0.1647 & -0.1053 & 0.1229 & -0.0104 & -0.0788 \\
GAT                   & 0.5343 & -4.3082 & 0.2031 & -0.0060 & -1.5445 & -0.0710 & 0.6904 & 0.4674 & 0.1446 \\
\bottomrule
\end{tabular}%
}
\end{table}

\begin{figure}[H]
    \centering
    \subfloat[\label{subfig:first}]{%
        \includegraphics[width=0.48\textwidth]{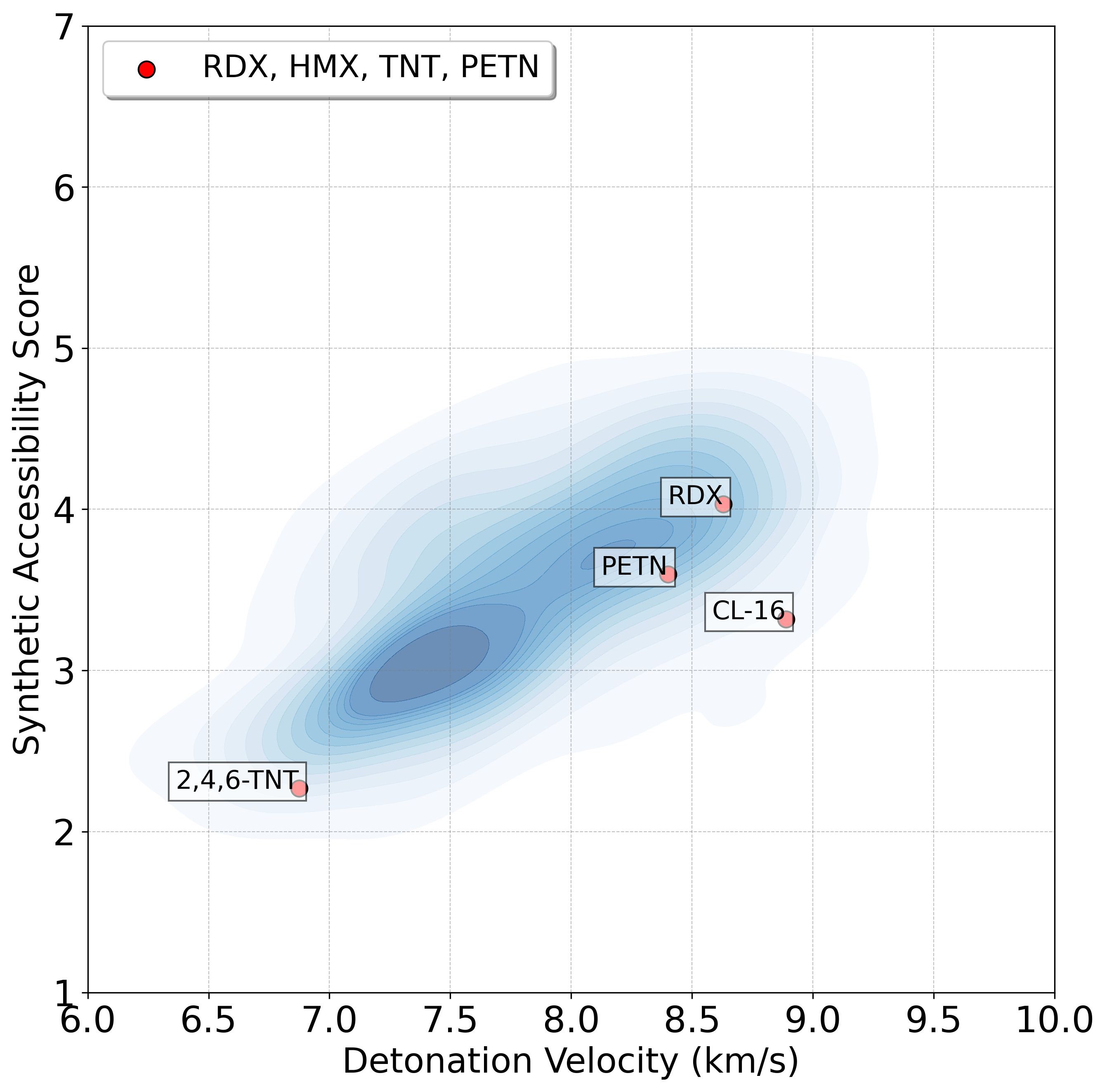}
    }
    \hfill
    \subfloat[\label{subfig:second}]{%
        \includegraphics[width=0.48\textwidth]{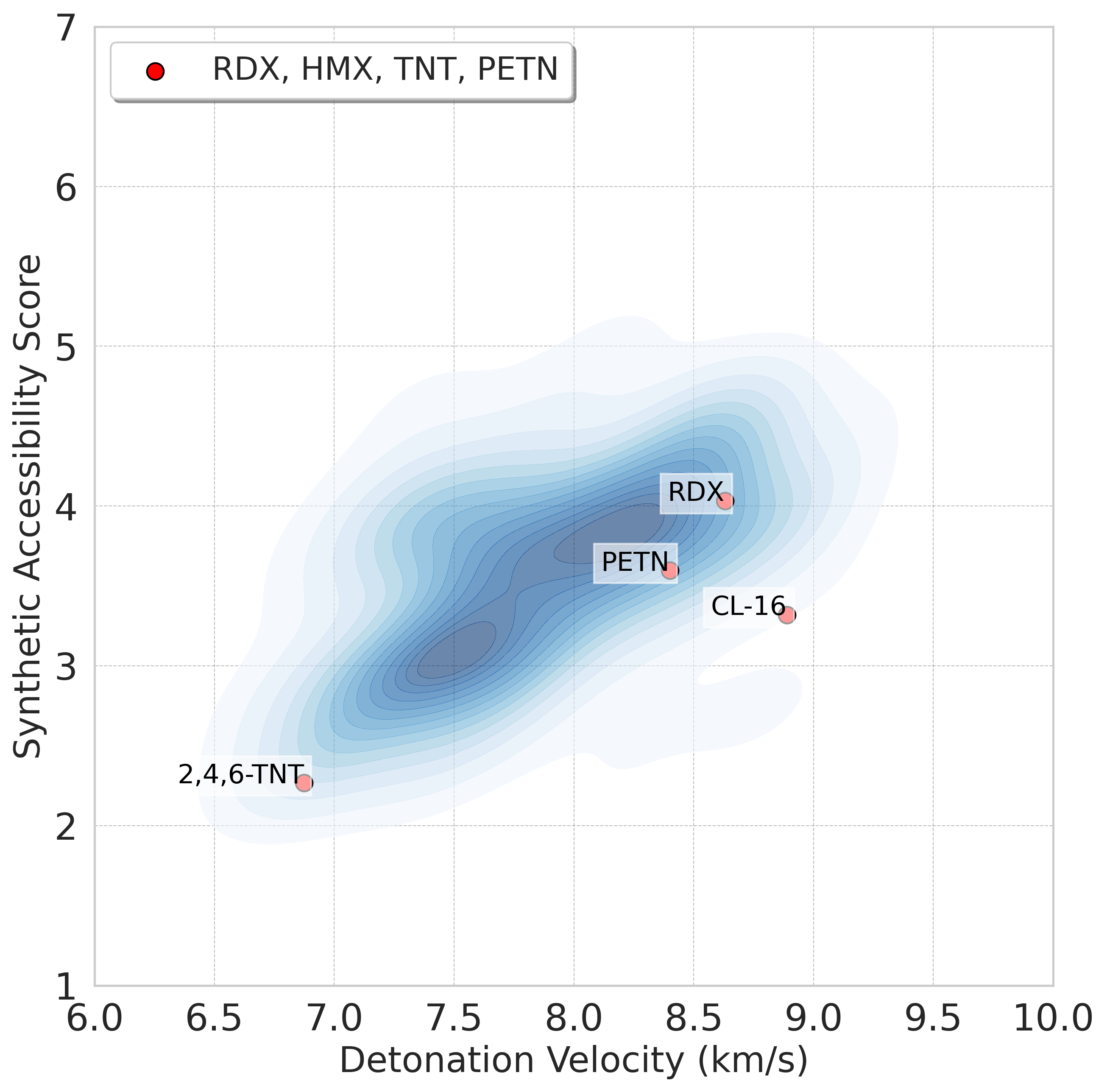}
    }
    \\
    \subfloat[\label{subfig:third}]{%
        \includegraphics[width=0.48\textwidth]{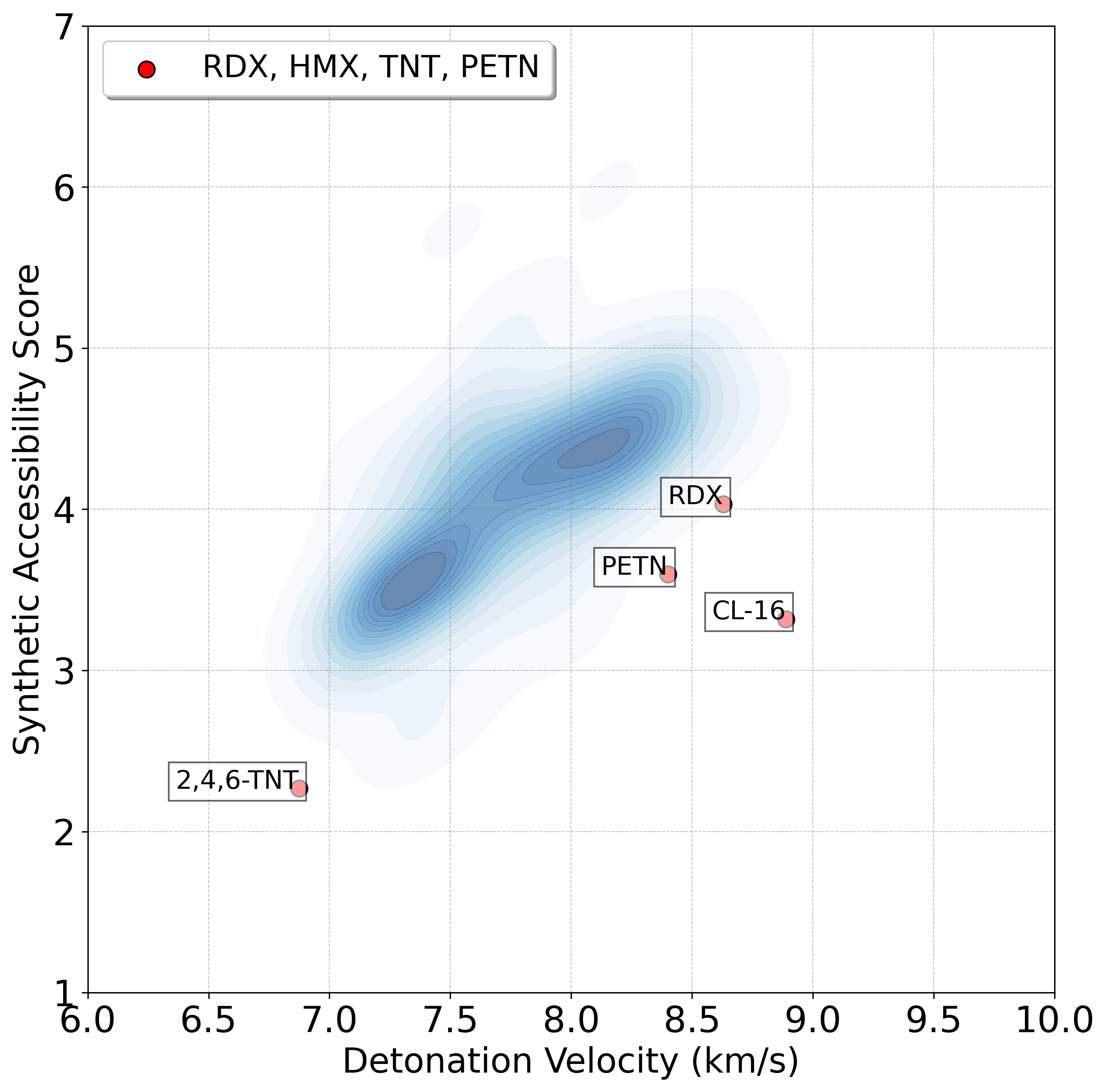}
    }
    \hfill
    \subfloat[\label{subfig:fourth}]{%
        \includegraphics[width=0.48\textwidth]{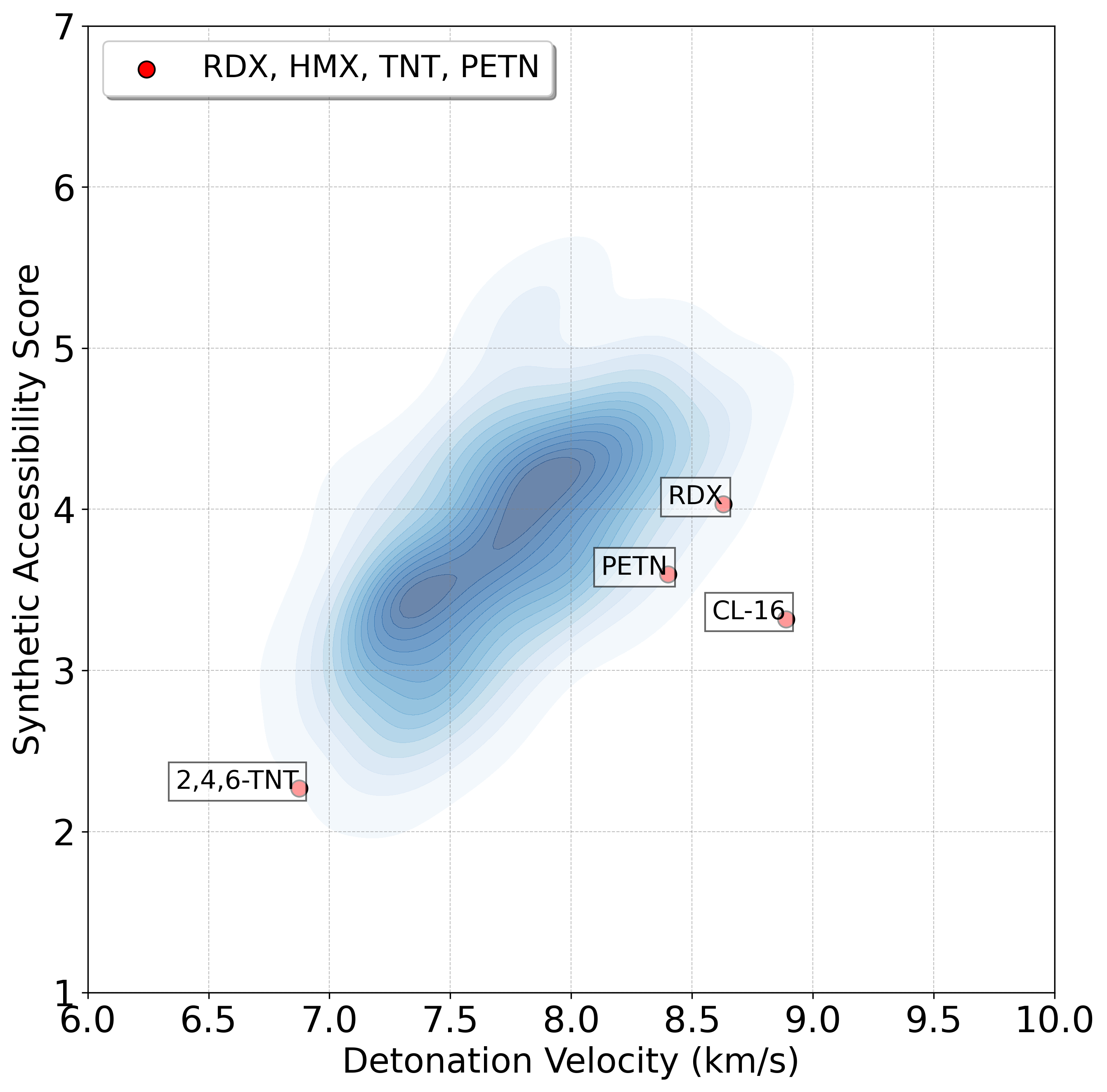}
    }
    \caption{Kernel Density Estimation (KDE) plots illustrating the relationship between the SA Score \cite{ertl2009estimation} and detonation velocity across different datasets and generative models.  
(a) KDE plot for the original dataset, serving as a baseline reference.  
(b) KDE plot for the generated dataset from a Simple LSTM model trained for 300 epochs with $3\times$ data augmentation, demonstrating the model’s ability to approximate the original distribution.  
(c) KDE plot for the generated dataset from the ULMFiT model, fine-tuned for 30 epochs on a $101\times$ augmented dataset, indicating an improved correlation structure.  
(d) KDE plot for the generated dataset from the ULMFiT model, fine-tuned for 40 epochs on a $101\times$ augmented dataset, showing further refinement in the predictive distribution. These visualizations provide insights into the fidelity of generated molecular properties compared to the original dataset. From these plots we are able to see that the region of generation of molecules of Simple LSTM is quite similar to the ULMFiT.}
    \label{fig:SA_SCORE_PLOTS}
\end{figure}

To evaluate the chemical design capabilities of different models, we have constructed a series of two dimensional kernel density plot as can be seen in Fig \ref{fig:SA_SCORE_PLOTS} mapping detonation velocity against Synthetic Accessibility Score (SAS). SAS, originally developed by Ertl and Schuffenhauer \cite{ertl2009estimation}, provides a quantitative estimate of a molecule’s ease of synthesis on a scale from 1 (very accessible) to 10 (highly complex and difficult to synthesize). This score combines contributions from molecular complexity—such as the presence of ring systems and stereo-centers with fragment-based frequency analysis derived from large chemical databases. By integrating SAS into our analysis, we aim to capture not only the energetic performance of generated molecules but also their synthetic feasibility, providing a more realistic assessment of each model’s utility for practical molecular design.

\begin{figure}[htbp]
    \centering
    \subfloat[Generation space of molecules from ULMFiT Model fine-tuned at 40 epochs on $101\times$ augmented dataset \label{subfig:ulmfit_gen}]{%
        \includegraphics[width=0.48\textwidth]{40_EPOCHULMFIT_SA_SCORE.png}
    }
    \hfill
    \subfloat[Generation space of molecules from Model 2 \label{subfig:model2_gen}]{%
        \includegraphics[width=0.48\textwidth]{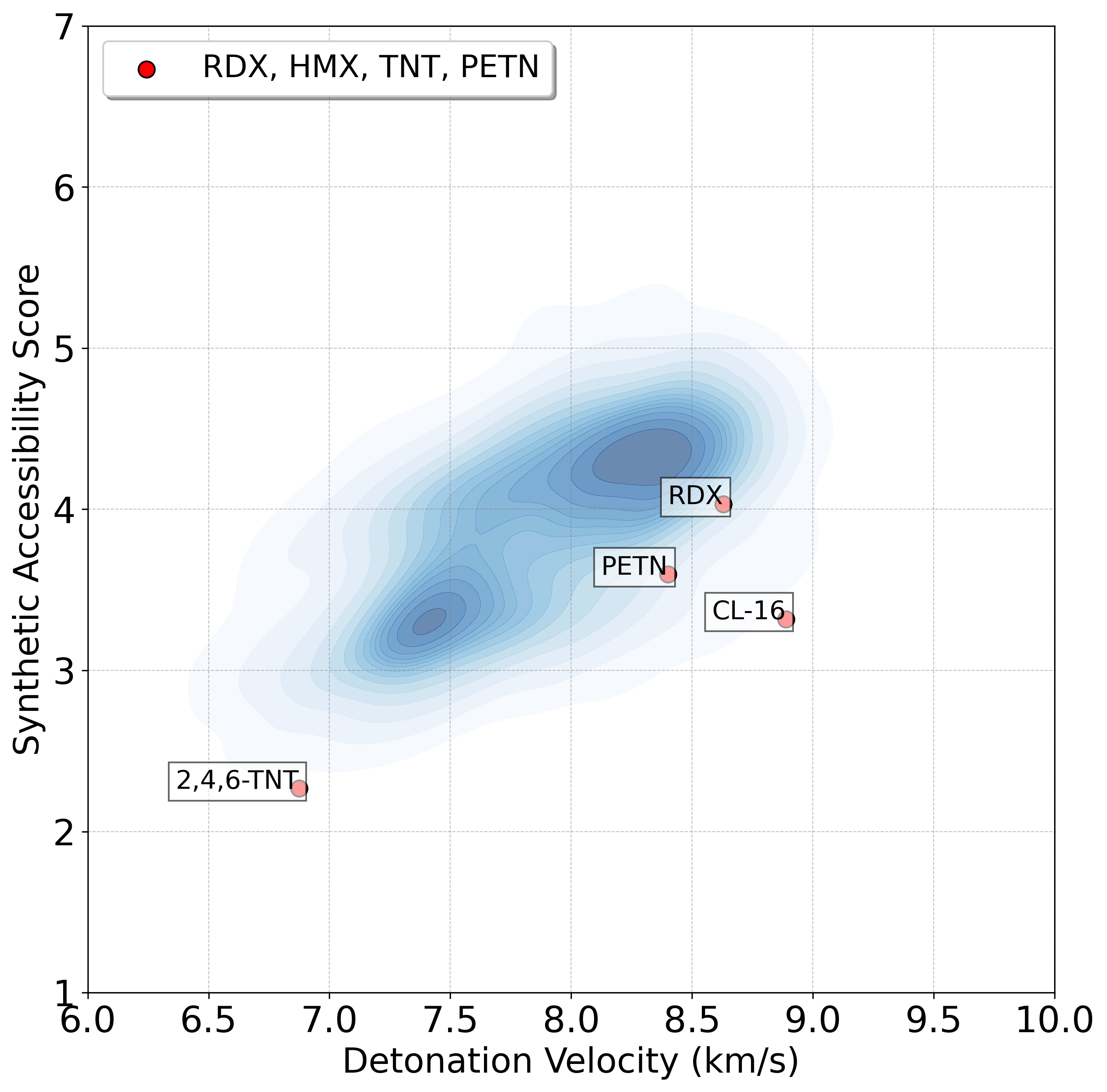}
    }
    \\
    \subfloat[Generation space of molecules from Model 3 \label{subfig:model3_gen}]{%
        \includegraphics[width=0.48\textwidth]{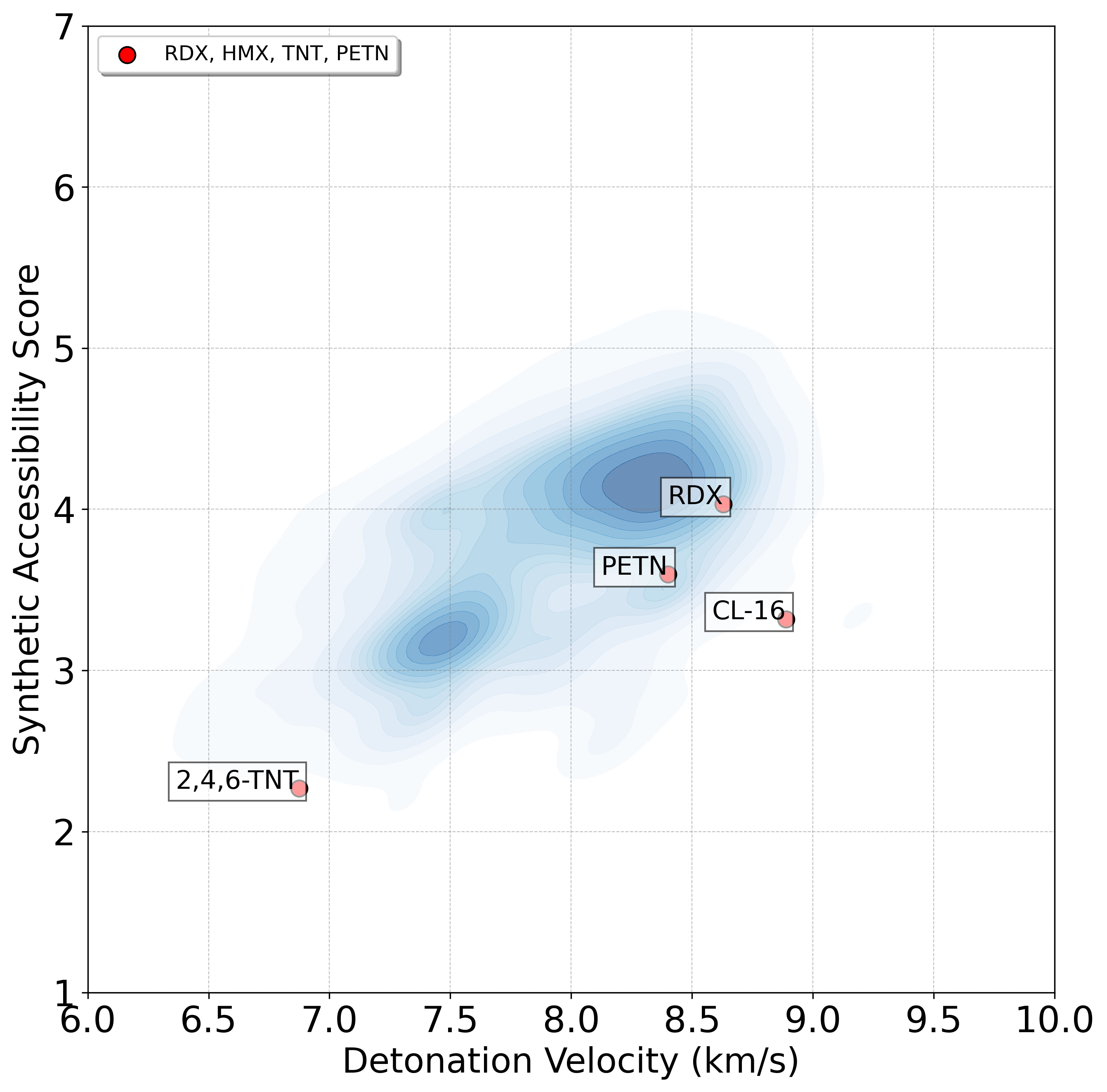}
    }
    \caption{Visualization of the generation space for different models.  
    (a) ULMFiT model fine-tuned for 40 epochs on a $101\times$ augmented dataset demonstrates a broad and diverse generation space, closely resembling the original data distribution.  
    (b) Model 2 exhibits a more constrained generation region, indicating potential mode collapse or limited generalization in molecular generation.  
    (c) Model 3 shows intermediate diversity and spread, balancing between the ULMFiT and Model 2 patterns. These plots collectively highlight how different generative architectures explore chemical space and the effectiveness of training strategies in achieving diverse molecule generation.}
    \label{fig:GENERATION_SPACE_PLOTS}
\end{figure}

Figure \ref{fig:SA_SCORE_PLOTS} and \ref{fig:GENERATION_SPACE_PLOTS} allow a direct assessment of the trade-off between a molecule's energetic performance and its feasibility of synthesis. In all the plots, we highlight well-known high-performance explosives- RDX, PETN and TNT- as anchor points for practical relevance. Figure \ref{subfig:first} shows the broad and balanced distribution of the original dataset we have used in training our models, This serves as reference baseline for evaluation. 
The ULMFiT-generated space as shown in Fig \ref{subfig:third} and \ref{subfig:fourth} shows a much tighter density concentrated around the high-performing and synthetically accessible region, particularly near RDX and PETN. This indicates that transfer learning from general chemical language followed by fine-tuning on the energetic compound domain, effectively narrows the generative space toward chemically meaningful and operationally relevant candidates. 

In contrast, the LSTM model molecular generation space as seen in Fig \ref{subfig:second} covers a similar region but with more noise and less density around the optimal region of real explosives. While this reflects a wider exploratory capacity, it also highlights limited inductive bias of the LSTM model and reduced ability to concentrate generation around practically useful compounds.

Despite ULMFiT showing stronger alignment with desirable regions, it exhibits a notable drawback: a lack of structural diversity. Its narrow focus may limit the discovery of novel, uncharted molecular scaffolds that lie outside the known performance landscape. This trade-off between precision and exploration is central to model selection for molecular generation. ULMFiT is well suited for exploitation of known chemical motifs, while LSTM models, though less precise, may be more effective in the early stages of discovery where chemical novelty is prioritized.


Figure \ref{fig:GENERATION_SPACE_PLOTS} shows a comparison of generation space across ULMFiT, Model 2, and Model 3 highlighting critical differences. Model 3 in particular demonstrates a improved generative behavior compared to earlier baselines, showing a concentrated and strategically skewed generation space toward molecules that are having high detonation velocity and low SA score. This shift in generative behavior can be directly attributed to a key architectural difference in embedding strategies of model.

While both Model 2 and Model 3 use partially trainable embeddings \(\mathbf{E} = [\mathbf{E}_{train}, \mathbf{E}_{fixed}]\), the difference in the initialization of the fixed component \(\mathbf{E}_{fixed}\) critically affects the model's learning dynamics and generative capability. In Model 2, \(\mathbf{E}_{fixed}\) is randomly initialized with Kaiming uniform distribution \cite{he2015delvingdeeprectifierssurpassing} as

\begin{equation}
\mathbf{E}_{\text{fixed}}^{(i)} \sim 
\mathcal{U}\!\left( 
    -\sqrt{\tfrac{6}{\mathrm{fan}_{\text{in}}}}, \; 
     \sqrt{\tfrac{6}{\mathrm{fan}_{\text{in}}}}
\right),
\end{equation}
where each token embedding vector $i$ is an independent random point. This randomness yields an embedding space lacking structured geometry, forcing the trainable component \(\mathbf{E}_{train}\) to approximate token identity mappings while also encoding property correlations. This dual task increases the hypothesis complexity and gradient variance, impairing convergence and leading to a less stable latent space representation.

Conversely, Model 3 constructs \(\mathbf{E}_{fixed}\) deterministically via SHA-256 hashing, as given by

\begin{equation}
\mathbf{E}_{\text{fixed}}^{(i)} 
= \operatorname{Normalize}\!\left( \operatorname{SHA256}( \text{token}_i ) \right),
\end{equation}
which provides a unique, fixed vector encoding each token’s identity with stable and reproducible structure. From a linear algebra perspective, these deterministic embeddings form a fixed orthonormal-like basis (or a low-coherence frame) in the input space, reducing embedding space entropy and conditioning the optimization landscape. Thus, the trainable embeddings $\mathbf{E}_{train}$ only need to learn semantic, structure-related transformations on top of this stable identity foundation, effectively lowering the intrinsic dimensionality of the learnable subspace as.

\begin{equation}
\min_{\mathbf{E}_{\text{train}}} \; 
\mathcal{L}\!\left( f\!\left(\mathbf{E}_{\text{train}} + \mathbf{E}_{\text{fixed}}\right), \, y \right),
\end{equation}
where \(f(\cdot)\) is the model mapping and \(y\) target properties.
This structured inductive bias improves gradient signal-to-noise ratio during backpropagation, leading to faster and more robust convergence. Consequently, Model 3 learns a more coherent latent representation \(\mathbf{z}\) that better aligns with molecular properties. The result is a tighter, more focused generation space that prioritizes molecules with optimized multi-objective properties, an outcome difficult to realize with the noisy, unstructured embeddings in Model 2.

\begin{table}[h!]
    \centering
    \caption{Average Tanimoto similarities (Morgan fingerprints) within and between datasets.}
    \label{tab:tanimoto_similarity}
    \begin{tabular}{l c}
        \toprule
        \textbf{Dataset Pair} & \textbf{Average Tanimoto Similarity} \\
        \midrule
        Intra-Dataset A (Original) & 0.2235 \\
        Intra-Dataset B (Model 2)  & 0.2061 \\
        Intra-Dataset C (Model 3)  & 0.2217 \\
        \midrule
        Inter-Dataset A vs B        & 0.2072 \\
        Inter-Dataset A vs C        & 0.2148 \\
        Inter-Dataset B vs C        & 0.2120 \\
        \bottomrule
    \end{tabular}
\end{table}
We computed average Tanimoto coefficients using Morgan fingerprints to evaluate the structural similarity within and across datasets. The intra-dataset similarity of the original dataset is 0.2235, indicating substantial chemical diversity. Model 3 achieves a comparable intra-similarity of 0.2217, closely preserving the original diversity, whereas Model 2 shows a slightly lower value of 0.2061, suggesting greater diversity but potentially less focused generation. Inter-dataset similarities reveal that Model 3 molecules share higher similarity with the original dataset (0.2148) compared to Model 2 (0.2072), demonstrating better alignment with the training distribution. The moderate similarity between Models 2 and 3 (0.2120) indicates partially overlapping yet distinct chemical spaces.

\begin{figure}[h!]
    \centering
    \includegraphics[width=0.9\textwidth]{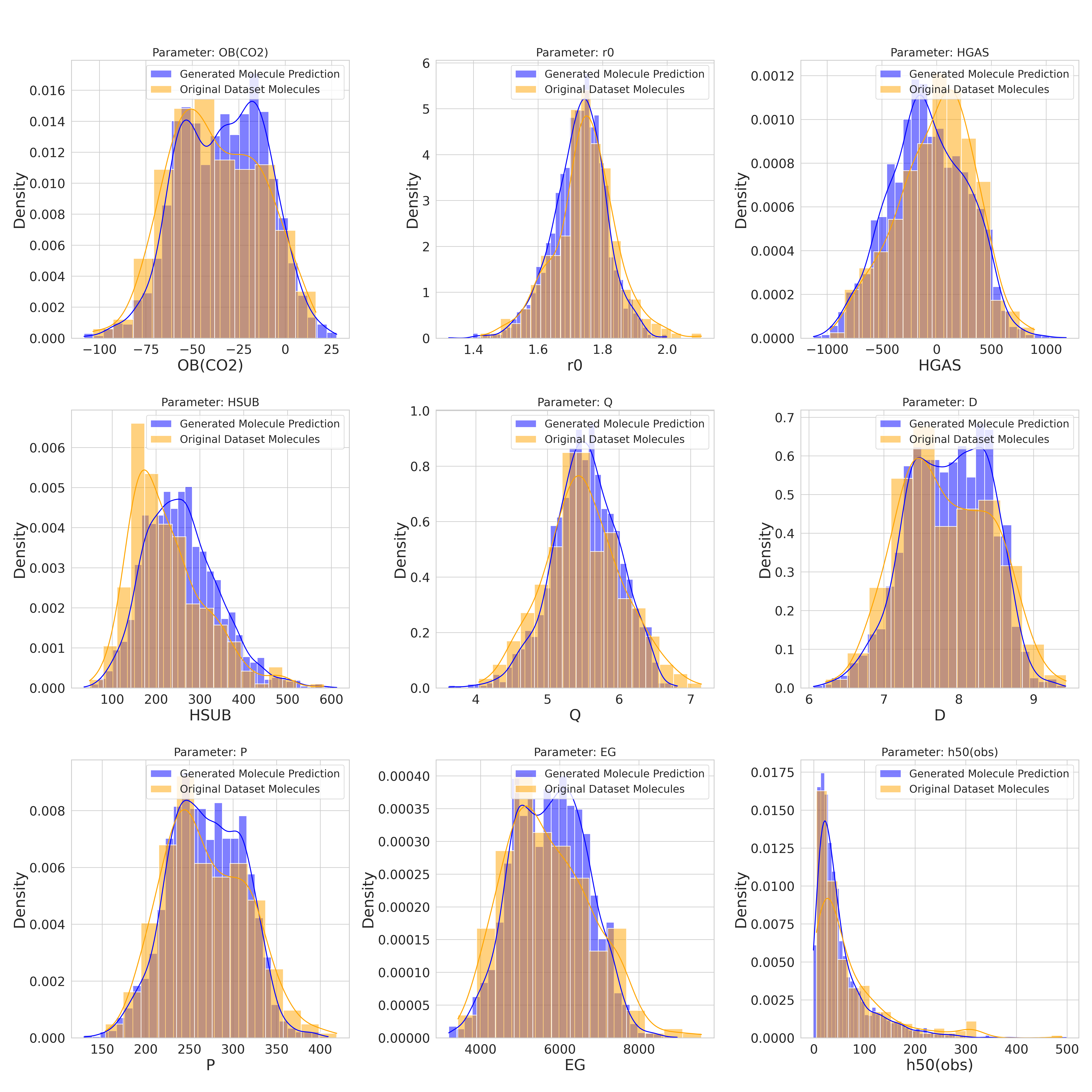}  
    \caption{Histogram comparison of properties of molecules generated from Model 2 and original dataset.}
    \label{model2histo}
\end{figure}
\begin{figure}[h!]
    \centering
    \includegraphics[width=0.9\textwidth]{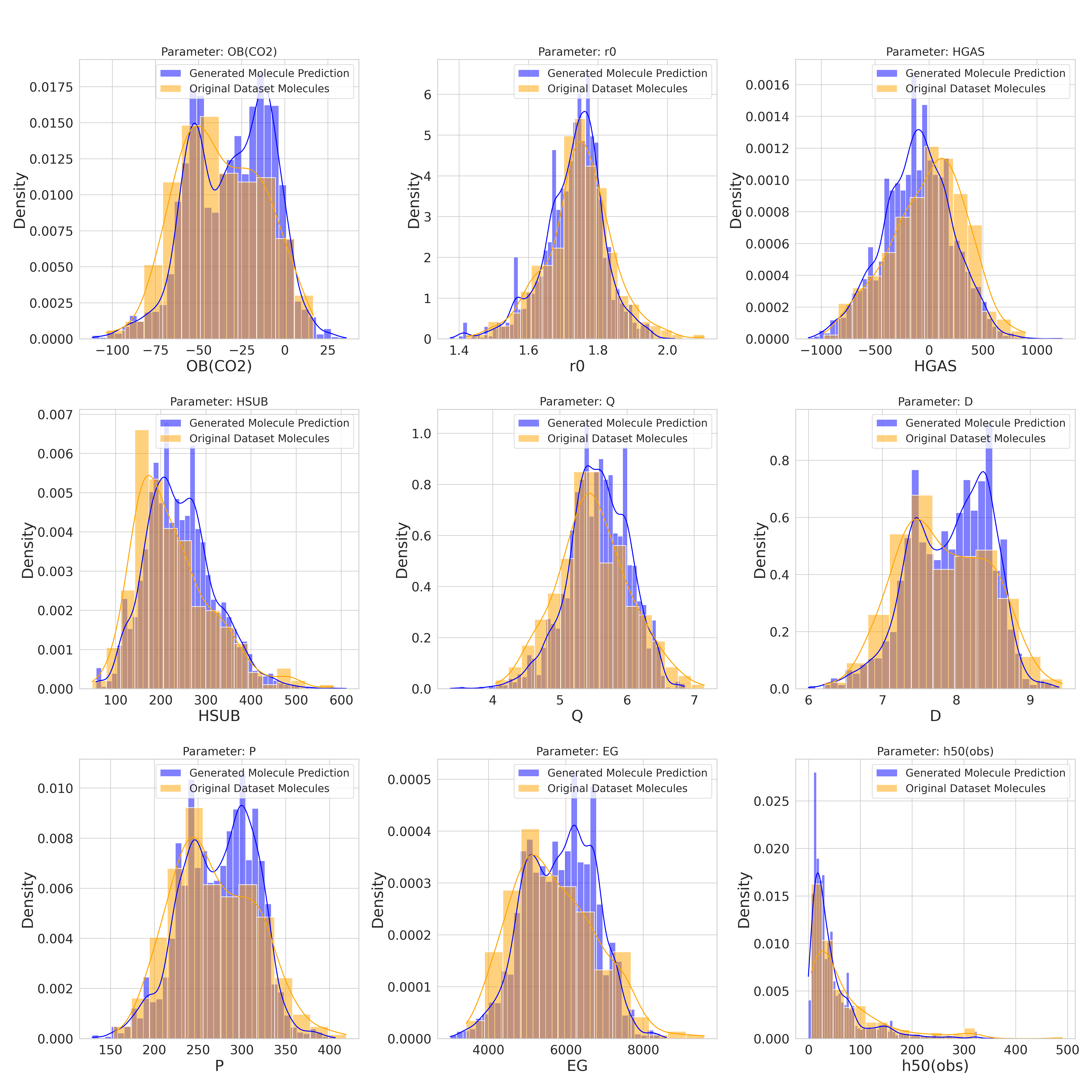}  
    \caption{Histogram comparison of properties of molecules generated from Model 3  and original dataset.}
    \label{model3histo}
\end{figure}

The LSTM--based generative approach yielded a chemically diverse set of energetic molecules, many exhibiting properties that align with state-of-the-art understanding of explosive sensitivity and performance. Upon critical analysis of predicted detonation velocity, pressure and oxygen balance distributions as shown in Fig. \ref{model2histo} and \ref{model3histo}, It is revealed that a significant fraction of compounds outperform legacy energetics in at least one critical metric.

\begin{figure}[htbp]
    \centering
    \includegraphics[width=0.95\textwidth]{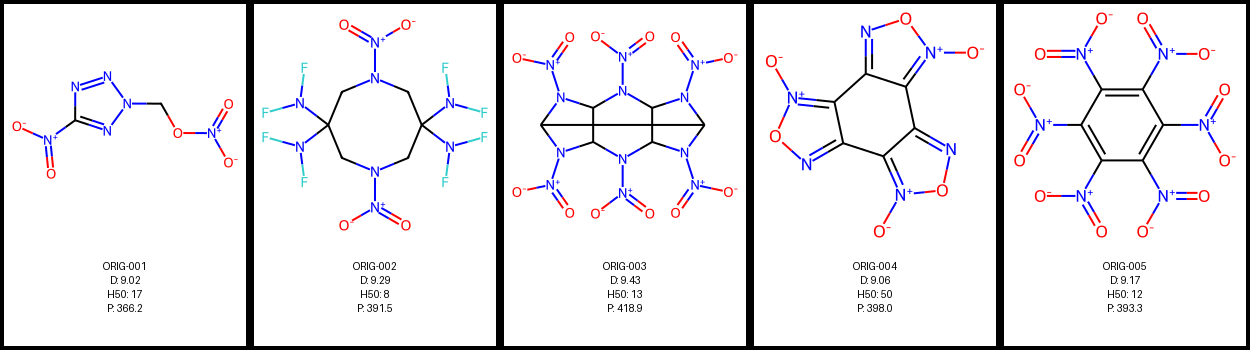}
    \caption{Top energetic molecules from the original dataset (\textit{D} $>$ 9 $km. s^{-1}$). The molecules are annotated with computed energetic properties such as detonation velocity (\textit{D}), detonation pressure (\textit{P}), and drop-weight impact sensitivity (h$_{50}$(obs)).}
    \label{fig:original}
\end{figure}

\begin{figure}[htbp]
    \centering
    \includegraphics[width=1\textwidth]{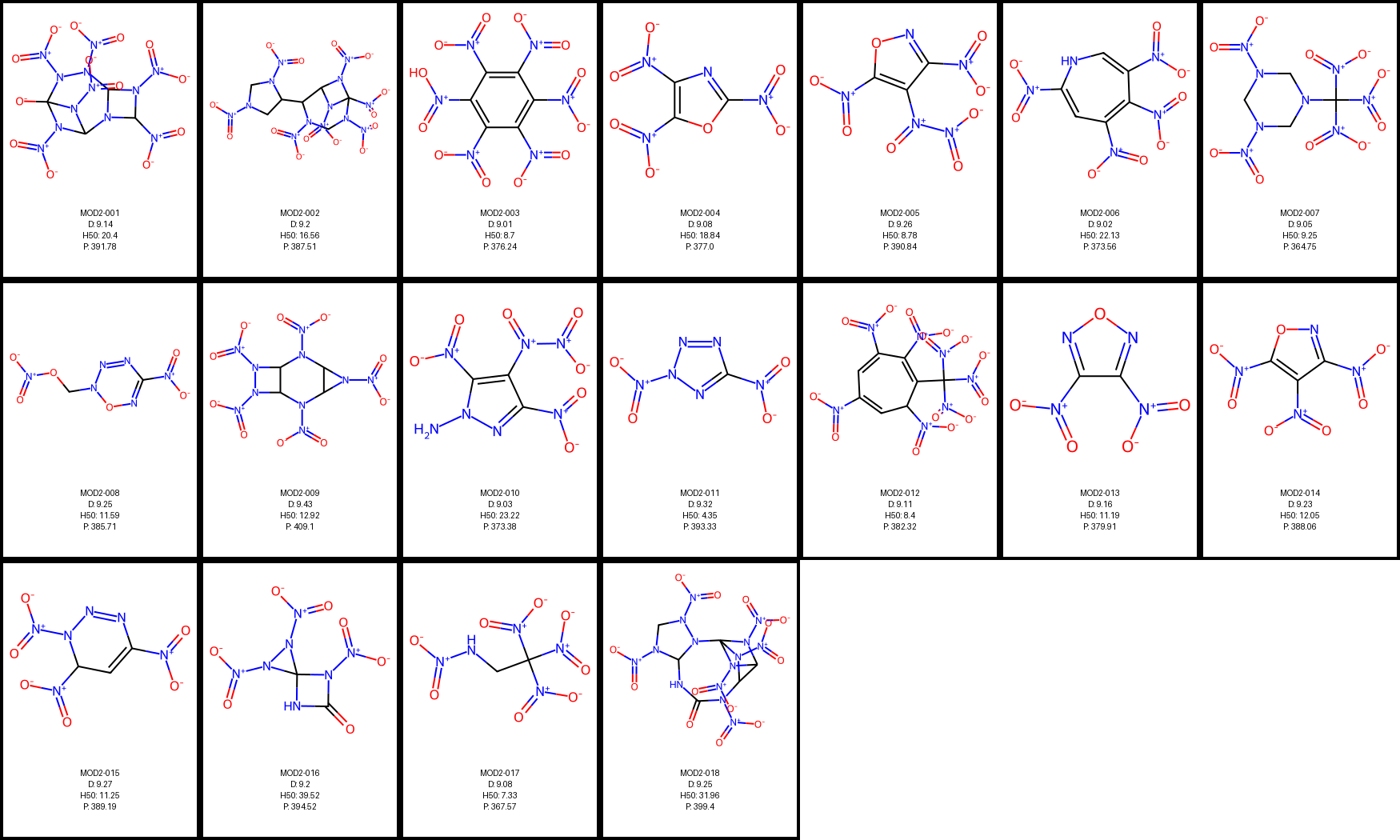}
    \caption{Top candidates generated by Model 2 exhibiting detonation velocity \textit{D} $>$ 9 $km. s^{-1}$. These molecules are also annotated with computed energetic properties (\textit{D}, \textit{P}, and h$_{50}$(obs)) to evaluate performance and sensitivity trade-offs.}
    \label{fig:model2}
\end{figure}

\begin{figure}[h!]
    \centering
    \includegraphics[width=1\textwidth]{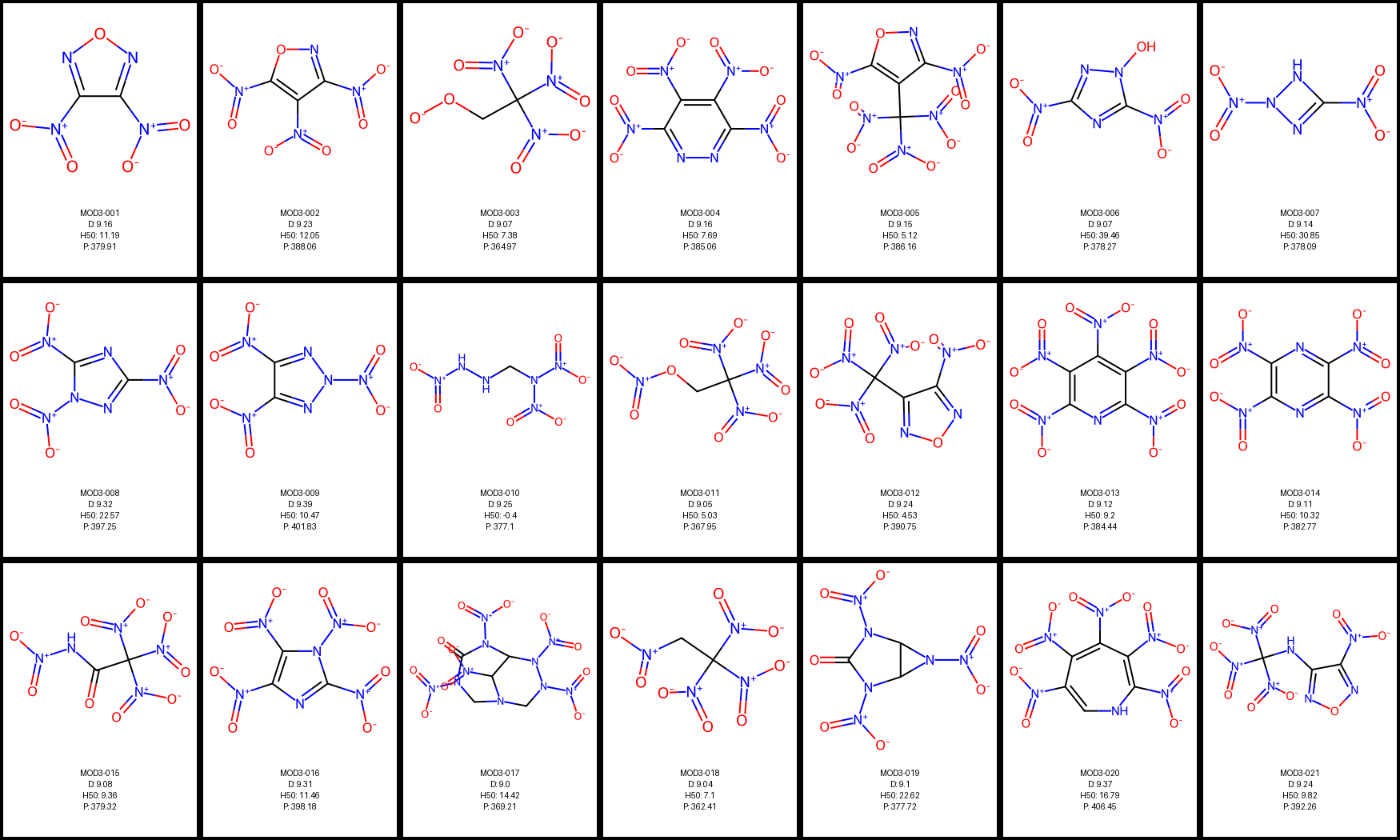}
    \caption{Top candidates generated by Model 3 with \textit{D} $>$ 9 $km. s^{-1}$. Molecular annotations include computed \textit{D}, \textit{P}, and h$_{50}$(obs) values, providing visual insight into their high-performance characteristics.}
    \label{fig:model3}
\end{figure}

 We conducted a focused analysis on compounds exhibiting high detonation velocity (\( D > 9 \)), a critical metric for evaluating high-energy materials performance \cite{doi:10.1021/acs.jpca.0c05969,doi:10.1021/acs.jpclett.2c02701, Cawkwell2022understanding}. Using RDKit, molecules from the original dataset, Model 2 and Model 3 were visualized and annotated with unique identifiers alongside detonation pressure (\( P \)) and experimental impact sensitivity (\( h_{50} \)) values as can be see in Figures \ref{fig:original} ,\ref{fig:model2} and \ref{fig:model3}.

We have designated each candidate in Fig \ref{fig:original}, \ref{fig:model2} and \ref{fig:model3} with certain abbreviation code indicating its source and identity. For example, MOD2-001 denotes that the molecule was generated using Model 2, with ‘001’ serving as its unique code.

Among the candidates, triazole- and tetrazole-based structures functionalized with nitrate ester and nitramine groups were prominent. This result directly supports findings by Li et al. \cite{li2022tri}, who demonstrated that fusing multiple explosophoric groups within nitrogen-rich heterocycles leads to superior energetic and safety profiles. These molecular-structures reflect design principles well-known in traditional energetic materials, yet the models extended and recombined them in novel, synthetically feasible arrangements.

Within MOD2 candidates (See Fig.\ref{fig:model2}), molecules such as MOD2-018 and MOD2-009 stand out for their fused or bridged ring systems incorporating multiple  groups. MOD2-018, in particular, exhibits a complex architecture that integrates nitramine and nitrate ester functionalities across a bridged polycyclic scaffold. This molecule achieves a detonation velocity of 9.25 $km. s^{-1}$ and pressure of 399.4 GPa, with a moderate predicted impact sensitivity of 31.96 cm, making it one of the most balanced high-performance candidates identified. Similarly, MOD2-009 achieves a detonation pressure over 400 GPa and exemplifies a compact, oxygen-rich structure.

Model 3 (See Fig.\ref{fig:model3}), shows incremental enhancement in the detonation velocity and higher stability. Compounds like MOD3-008, MOD3-009, and MOD3-016 exhibit high detonation velocities above 9.3 $km. s^{-1}$ and pressures nearing or exceeding 400 GPa, yet maintain lower predicted sensitivities compared to their MOD2 counterparts. These molecules typically feature fused nitrogen-rich rings, symmetric substitution patterns, and minimal ring strain, indicating that the model has internalized fundamental chemical heuristics relevant to energetic compound design. For instance, MOD3-020 offers a high-performance profile with a detonation velocity of 9.37 $km. s^{-1}$ and an impact sensitivity of 16.79 cm, positioning it as an excellent candidate for experimental synthesis. Similarly, MOD3-006 and MOD3-012 exemplify how structurally clean azole frameworks with appropriate explosophore substitutions can simultaneously meet detonation, safety, and feasibility criteria.

Despite being trained independently, both MOD2 and MOD3 converge on a set of structural motifs that appear repeatedly among the top candidates. Azole based cores are central to many high-performing molecules due to their high nitrogen content and versatility for substitution. Nitrate esters and nitramines are the most common functional groups attached to these heterocycles, consistent with their role in boosting oxygen balance and energy release. Additionally, fused or bicyclic frameworks frequently emerge in high-ranking structures, as they contribute to higher density, better oxygen balance, and a favorable detonation-to-sensitivity trade-off. Compared to compounds seen in the original dataset, which often relied on extensively nitrated polycyclic systems, the generated molecules favor more compact, synthetically plausible designs while still achieving equivalent or superior energetic metrics.

Analysis of the overall property distributions confirms the generative model success in learning useful structure-property relationships. Model 3, in particular, displays a narrower distribution in impact sensitivity while achieving higher average detonation velocities and pressures compared to Model 2. This suggests that the architecture improvements and learning mechanisms in Model 3 enable a more focused search in the chemical space of viable energetic materials. In contrast, Model 2 explores a broader and more diverse structural space, surfacing creative and novel topologies, some of which push the limits of chemical feasibility. This distinction positions Model 3 as the preferred model for high-throughput lead identification, while Model 2 remains valuable for ideation and expanding the design space.

Overall, the results highlight the effectiveness of our generative approach in producing not only high-performing but also chemically reasonable energetic molecules. The most promising candidates—such as MOD3-006, MOD3-007, MOD2-001 and MOD2-018—demonstrate that the models can generate novel compounds that match or exceed traditional materials in detonation velocity and pressure, while maintaining moderate sensitivities and plausible synthetic accessibility. The emergence of triazole- and tetrazole-based frameworks with well-placed nitrate ester and nitramine groups suggests a clear direction for experimental efforts. Moreover, the fusion of these motifs into structurally compact and symmetric systems offers a path toward designing next-generation high-energy compounds with improved performance-safety profiles.

\section{Conclusion}

The study of high-energy molecules and compounds have been a very restricted area thereby very--limited public data has been available on these topic, making it a challenging problem to work with low--data and restricted infrastructure for new researchers to work independently in this field. Development and Production of such high--energy molecules parallels the military research activities across globe, accelerating the work in design of high--energy molecules using machine learning practices could provide a huge leap to research in this field.

Through our work we have introduced a novel, lightweight generative framework that enables the discovery of high-energy molecular candidates under extreme data scarcity and computational limitations. By leveraging a simple yet effective LSTM-based generative model with partially trainable SHA-256 derived embedding and combining it with AttentiveFP, an Attentive Graph Neural Network for property prediction, our approach not only matches but in some cases surpasses the generative and predictive performance of models built with orders of magnitude more data and resources.

We have created a database of 4900+ new candidate molecules, several of which exhibit detonation velocities exceeding 9$km. s^{-1}$,comparable or superior to legacy materials such as RDX and PETN. We have shown that our generative architecture is capable of learning rich structure-property relationships and producing novel chemical motifs, including functionalized triazole and tetrazole scaffolds that have practically high energetic potential.

Notably, our study underscores the viability of pursuing high-performance material discovery without reliance on pretraining or large-scale chemical corpora thereby opening the doors for independent research groups and academic laboratories to make impactful contributions using only publicly available data and commodity hardware. Our framework demonstrates that informed design decisions such as embedding strategy and domain-attuned learning objectives can mitigate data scarcity challenges.

By distilling a complex generative task into a fast, explainable and low-footprint architecture, we have taken the first step toward building autonomous scientific assistants—models that not only discover but understand, filter and align chemical innovations with broader goals.

Beyond its technical novelty, this work reveals the dual-use nature of scientific progress. While our methodology can accelerate the development of cleaner propellants and safer explosives for space missions or mining applications, it equally highlights the potential for misuse. The ability to generate novel energetic compounds at scale—even with publicly available tools—raises ethical questions about access, intent, and governance. As AI capabilities grow more autonomous, so too must our vigilance in aligning them with transparent, responsible frameworks.


\section{Acknowledgement}
\noindent AA and SV are grateful to Prof. Sunoj B. at IIT Bombay for fruitful discussions and suggestions without which this work would have not reached this state.

\section*{Data availability}
\noindent The data used in this work is available in literature already.





\bibliographystyle{elsarticle-num}
\bibliography{example}

\begin{thebibliography}{10}
\expandafter\ifx\csname url\endcsname\relax
  \def\url#1{\texttt{#1}}\fi
\expandafter\ifx\csname urlprefix\endcsname\relax\def\urlprefix{URL }\fi
\expandafter\ifx\csname href\endcsname\relax
  \def\href#1#2{#2} \def\path#1{#1}\fi

\bibitem{HUTCHINGS201972}
M.~I. Hutchings, A.~W. Truman, B.~Wilkinson,
  \href{https://www.sciencedirect.com/science/article/pii/S1369527419300190}{Antibiotics:
  past, present and future}, Current Opinion in Microbiology 51 (2019) 72--80,
  antimicrobials.
\newblock \href {https://doi.org/https://doi.org/10.1016/j.mib.2019.10.008}
  {\path{doi:https://doi.org/10.1016/j.mib.2019.10.008}}.
\newline\urlprefix\url{https://www.sciencedirect.com/science/article/pii/S1369527419300190}

\bibitem{agrawal2010high}
J.~Agrawal, High Energy Materials: Propellants Explosives and Pyrotechnics,
  John Wiley \& Sons, 2010.

\bibitem{doi:10.1021/acsomega.4c01070}
M.~Rachamim, A.~J. Domb, A.~Goldblum,
  \href{https://doi.org/10.1021/acsomega.4c01070}{Modeling high energy
  molecules and screening to find novel high energy candidates}, ACS Omega
  9~(42) (2024) 42709--42720.
\newblock \href {http://arxiv.org/abs/https://doi.org/10.1021/acsomega.4c01070}
  {\path{arXiv:https://doi.org/10.1021/acsomega.4c01070}}, \href
  {https://doi.org/10.1021/acsomega.4c01070}
  {\path{doi:10.1021/acsomega.4c01070}}.
\newline\urlprefix\url{https://doi.org/10.1021/acsomega.4c01070}

\bibitem{HUANG2021102240}
X.~Huang, C.~Li, K.~Tan, Y.~Wen, F.~Guo, M.~Li, Y.~Huang, C.~Q. Sun, M.~Gozin,
  L.~Zhang,
  \href{https://www.sciencedirect.com/science/article/pii/S258900422100208X}{Applying
  machine learning to balance performance and stability of high energy density
  materials}, iScience 24~(3) (2021) 102240.
\newblock \href {https://doi.org/https://doi.org/10.1016/j.isci.2021.102240}
  {\path{doi:https://doi.org/10.1016/j.isci.2021.102240}}.
\newline\urlprefix\url{https://www.sciencedirect.com/science/article/pii/S258900422100208X}

\bibitem{MEYERS20212707}
J.~Meyers, B.~Fabian, N.~Brown,
  \href{https://www.sciencedirect.com/science/article/pii/S1359644621002531}{De
  novo molecular design and generative models}, Drug Discovery Today 26~(11)
  (2021) 2707--2715.
\newblock \href {https://doi.org/https://doi.org/10.1016/j.drudis.2021.05.019}
  {\path{doi:https://doi.org/10.1016/j.drudis.2021.05.019}}.
\newline\urlprefix\url{https://www.sciencedirect.com/science/article/pii/S1359644621002531}

\bibitem{doi:10.1126/science.aat2663}
B.~Sanchez-Lengeling, A.~Aspuru-Guzik,
  \href{https://www.science.org/doi/abs/10.1126/science.aat2663}{Inverse
  molecular design using machine learning: Generative models for matter
  engineering}, Science 361~(6400) (2018) 360--365.
\newblock \href
  {http://arxiv.org/abs/https://www.science.org/doi/pdf/10.1126/science.aat2663}
  {\path{arXiv:https://www.science.org/doi/pdf/10.1126/science.aat2663}}, \href
  {https://doi.org/10.1126/science.aat2663}
  {\path{doi:10.1126/science.aat2663}}.
\newline\urlprefix\url{https://www.science.org/doi/abs/10.1126/science.aat2663}

\bibitem{BAILLIF2023102566}
B.~Baillif, J.~Cole, P.~McCabe, A.~Bender,
  \href{https://www.sciencedirect.com/science/article/pii/S0959440X23000404}{Deep
  generative models for 3d molecular structure}, Current Opinion in Structural
  Biology 80 (2023) 102566.
\newblock \href {https://doi.org/https://doi.org/10.1016/j.sbi.2023.102566}
  {\path{doi:https://doi.org/10.1016/j.sbi.2023.102566}}.
\newline\urlprefix\url{https://www.sciencedirect.com/science/article/pii/S0959440X23000404}

\bibitem{gaudelet2021utilizing}
T.~Gaudelet, B.~Day, A.~R. Jamasb, J.~Soman, C.~Regep, G.~Liu, J.~B. Hayter,
  R.~Vickers, C.~Roberts, J.~Tang, et~al., Utilizing graph machine learning
  within drug discovery and development, Briefings in bioinformatics 22~(6)
  (2021) bbab159.

\bibitem{GRISONI2023102527}
F.~Grisoni,
  \href{https://www.sciencedirect.com/science/article/pii/S0959440X23000015}{Chemical
  language models for de novo drug design: Challenges and opportunities},
  Current Opinion in Structural Biology 79 (2023) 102527.
\newblock \href {https://doi.org/https://doi.org/10.1016/j.sbi.2023.102527}
  {\path{doi:https://doi.org/10.1016/j.sbi.2023.102527}}.
\newline\urlprefix\url{https://www.sciencedirect.com/science/article/pii/S0959440X23000015}

\bibitem{HANSER2023102545}
T.~Hanser,
  \href{https://www.sciencedirect.com/science/article/pii/S0959440X23000192}{Federated
  learning for molecular discovery}, Current Opinion in Structural Biology 79
  (2023) 102545.
\newblock \href {https://doi.org/https://doi.org/10.1016/j.sbi.2023.102545}
  {\path{doi:https://doi.org/10.1016/j.sbi.2023.102545}}.
\newline\urlprefix\url{https://www.sciencedirect.com/science/article/pii/S0959440X23000192}

\bibitem{wang2023scientific}
H.~Wang, T.~Fu, Y.~Du, W.~Gao, K.~Huang, Z.~Liu, P.~Chandak, S.~Liu,
  P.~Van~Katwyk, A.~Deac, et~al., Scientific discovery in the age of artificial
  intelligence, Nature 620~(7972) (2023) 47--60.

\bibitem{ISERT2023102548}
C.~Isert, K.~Atz, G.~Schneider,
  \href{https://www.sciencedirect.com/science/article/pii/S0959440X23000222}{Structure-based
  drug design with geometric deep learning}, Current Opinion in Structural
  Biology 79 (2023) 102548.
\newblock \href {https://doi.org/https://doi.org/10.1016/j.sbi.2023.102548}
  {\path{doi:https://doi.org/10.1016/j.sbi.2023.102548}}.
\newline\urlprefix\url{https://www.sciencedirect.com/science/article/pii/S0959440X23000222}

\bibitem{THOMAS2023102559}
M.~Thomas, A.~Bender, C.~{de Graaf},
  \href{https://www.sciencedirect.com/science/article/pii/S0959440X23000337}{Integrating
  structure-based approaches in generative molecular design}, Current Opinion
  in Structural Biology 79 (2023) 102559.
\newblock \href {https://doi.org/https://doi.org/10.1016/j.sbi.2023.102559}
  {\path{doi:https://doi.org/10.1016/j.sbi.2023.102559}}.
\newline\urlprefix\url{https://www.sciencedirect.com/science/article/pii/S0959440X23000337}

\bibitem{zhang2023equivariant}
Z.~Zhang, Q.~Liu, C.-K. Lee, C.-Y. Hsieh, E.~Chen, An equivariant generative
  framework for molecular graph-structure co-design, Chemical Science 14~(31)
  (2023) 8380--8392.

\bibitem{doi:10.1021/ci100050t}
D.~Rogers, M.~Hahn,
  \href{https://doi.org/10.1021/ci100050t}{Extended-connectivity fingerprints},
  Journal of Chemical Information and Modeling 50~(5) (2010) 742--754, pMID:
  20426451.
\newblock \href {http://arxiv.org/abs/https://doi.org/10.1021/ci100050t}
  {\path{arXiv:https://doi.org/10.1021/ci100050t}}, \href
  {https://doi.org/10.1021/ci100050t} {\path{doi:10.1021/ci100050t}}.
\newline\urlprefix\url{https://doi.org/10.1021/ci100050t}

\bibitem{zeng2022accurate}
X.~Zeng, H.~Xiang, L.~Yu, J.~Wang, K.~Li, R.~Nussinov, F.~Cheng, Accurate
  prediction of molecular properties and drug targets using a self-supervised
  image representation learning framework, Nature Machine Intelligence 4~(11)
  (2022) 1004--1016.

\bibitem{sherstinsky2020fundamentals}
A.~Sherstinsky, Fundamentals of recurrent neural network (rnn) and long
  short-term memory (lstm) network, Physica D: Nonlinear Phenomena 404 (2020)
  132306.

\bibitem{kingma2013auto}
D.~P. Kingma, Auto-encoding variational bayes, arXiv preprint arXiv:1312.6114
  (2013).

\bibitem{goodfellow2020generative}
I.~Goodfellow, J.~Pouget-Abadie, M.~Mirza, B.~Xu, D.~Warde-Farley, S.~Ozair,
  A.~Courville, Y.~Bengio, Generative adversarial networks, Communications of
  the ACM 63~(11) (2020) 139--144.

\bibitem{makhzani2015adversarial}
A.~Makhzani, J.~Shlens, N.~Jaitly, I.~Goodfellow, B.~Frey, Adversarial
  autoencoders, arXiv preprint arXiv:1511.05644 (2015).

\bibitem{martinelli2022generative}
D.~D. Martinelli, Generative machine learning for de novo drug discovery: A
  systematic review, Computers in Biology and Medicine 145 (2022) 105403.

\bibitem{olivecrona2017molecular}
M.~Olivecrona, T.~Blaschke, O.~Engkvist, H.~Chen, Molecular de-novo design
  through deep reinforcement learning, Journal of cheminformatics 9 (2017)
  1--14.

\bibitem{gomez2018automatic}
R.~G{\'o}mez-Bombarelli, J.~N. Wei, D.~Duvenaud, J.~M. Hern{\'a}ndez-Lobato,
  B.~S{\'a}nchez-Lengeling, D.~Sheberla, J.~Aguilera-Iparraguirre, T.~D.
  Hirzel, R.~P. Adams, A.~Aspuru-Guzik, Automatic chemical design using a
  data-driven continuous representation of molecules, ACS central science 4~(2)
  (2018) 268--276.

\bibitem{kadurin2016cornucopia}
A.~Kadurin, A.~Aliper, A.~Kazennov, P.~Mamoshina, Q.~Vanhaelen, K.~Khrabrov,
  A.~Zhavoronkov, The cornucopia of meaningful leads: Applying deep adversarial
  autoencoders for new molecule development in oncology, Oncotarget 8~(7)
  (2016) 10883.

\bibitem{polykovskiy2018entangled}
D.~Polykovskiy, A.~Zhebrak, D.~Vetrov, Y.~Ivanenkov, V.~Aladinskiy,
  P.~Mamoshina, M.~Bozdaganyan, A.~Aliper, A.~Zhavoronkov, A.~Kadurin,
  Entangled conditional adversarial autoencoder for de novo drug discovery,
  Molecular pharmaceutics 15~(10) (2018) 4398--4405.

\bibitem{Yang31122017}
X.~Yang, J.~Zhang, K.~Yoshizoe, K.~Terayama, K.~Tsuda,
  \href{http://dx.doi.org/10.1080/14686996.2017.1401424}{Chemts: an efficient
  python library for de novo molecular generation}, Science and Technology of
  Advanced Materials 18~(1) (2017) 972–976.
\newblock \href {https://doi.org/10.1080/14686996.2017.1401424}
  {\path{doi:10.1080/14686996.2017.1401424}}.
\newline\urlprefix\url{http://dx.doi.org/10.1080/14686996.2017.1401424}

\bibitem{van2020gen}
R.~van Deursen, P.~Ertl, I.~V. Tetko, G.~Godin, Gen: highly efficient smiles
  explorer using autodidactic generative examination networks, Journal of
  Cheminformatics 12 (2020) 1--14.

\bibitem{moret2020generative}
M.~Moret, L.~Friedrich, F.~Grisoni, D.~Merk, G.~Schneider, Generative molecular
  design in low data regimes, Nature Machine Intelligence 2~(3) (2020)
  171--180.

\bibitem{doi:10.1021/acs.jcim.9b00943}
F.~Grisoni, M.~Moret, R.~Lingwood, G.~Schneider,
  \href{https://doi.org/10.1021/acs.jcim.9b00943}{Bidirectional molecule
  generation with recurrent neural networks}, Journal of Chemical Information
  and Modeling 60~(3) (2020) 1175--1183, pMID: 31904964.
\newblock \href {http://arxiv.org/abs/https://doi.org/10.1021/acs.jcim.9b00943}
  {\path{arXiv:https://doi.org/10.1021/acs.jcim.9b00943}}, \href
  {https://doi.org/10.1021/acs.jcim.9b00943}
  {\path{doi:10.1021/acs.jcim.9b00943}}.
\newline\urlprefix\url{https://doi.org/10.1021/acs.jcim.9b00943}

\bibitem{https://doi.org/10.1002/minf.201700153}
D.~Merk, L.~Friedrich, F.~Grisoni, G.~Schneider,
  \href{https://onlinelibrary.wiley.com/doi/abs/10.1002/minf.201700153}{De novo
  design of bioactive small molecules by artificial intelligence}, Molecular
  Informatics 37~(1-2) (2018) 1700153.
\newblock \href
  {http://arxiv.org/abs/https://onlinelibrary.wiley.com/doi/pdf/10.1002/minf.201700153}
  {\path{arXiv:https://onlinelibrary.wiley.com/doi/pdf/10.1002/minf.201700153}},
  \href {https://doi.org/https://doi.org/10.1002/minf.201700153}
  {\path{doi:https://doi.org/10.1002/minf.201700153}}.
\newline\urlprefix\url{https://onlinelibrary.wiley.com/doi/abs/10.1002/minf.201700153}

\bibitem{wu2020comprehensive}
Z.~Wu, S.~Pan, F.~Chen, G.~Long, C.~Zhang, S.~Y. Philip, A comprehensive survey
  on graph neural networks, IEEE transactions on neural networks and learning
  systems 32~(1) (2020) 4--24.

\bibitem{jiang2021could}
D.~Jiang, Z.~Wu, C.-Y. Hsieh, G.~Chen, B.~Liao, Z.~Wang, C.~Shen, D.~Cao,
  J.~Wu, T.~Hou, Could graph neural networks learn better molecular
  representation for drug discovery? a comparison study of descriptor-based and
  graph-based models, Journal of cheminformatics 13 (2021) 1--23.

\bibitem{Rittig_2023}
J.~G. Rittig, Q.~Gao, M.~Dahmen, A.~Mitsos, A.~M. Schweidtmann,
  \href{http://dx.doi.org/10.1039/BK9781837670178-00159}{Graph Neural Networks
  for the Prediction of Molecular Structure–Property Relationships}, Royal
  Society of Chemistry, 2023, p. 159–181.
\newblock \href {https://doi.org/10.1039/bk9781837670178-00159}
  {\path{doi:10.1039/bk9781837670178-00159}}.
\newline\urlprefix\url{http://dx.doi.org/10.1039/BK9781837670178-00159}

\bibitem{doi:10.1021/acsomega.2c06702}
W.~Ahmad, H.~Tayara, K.~T. Chong,
  \href{https://doi.org/10.1021/acsomega.2c06702}{Attention-based graph neural
  network for molecular solubility prediction}, ACS Omega 8~(3) (2023)
  3236--3244.
\newblock \href {http://arxiv.org/abs/https://doi.org/10.1021/acsomega.2c06702}
  {\path{arXiv:https://doi.org/10.1021/acsomega.2c06702}}, \href
  {https://doi.org/10.1021/acsomega.2c06702}
  {\path{doi:10.1021/acsomega.2c06702}}.
\newline\urlprefix\url{https://doi.org/10.1021/acsomega.2c06702}

\bibitem{doi:10.1021/acs.jmedchem.9b00959}
Z.~Xiong, D.~Wang, X.~Liu, F.~Zhong, X.~Wan, X.~Li, Z.~Li, X.~Luo, K.~Chen,
  H.~Jiang, M.~Zheng,
  \href{https://doi.org/10.1021/acs.jmedchem.9b00959}{Pushing the boundaries of
  molecular representation for drug discovery with the graph attention
  mechanism}, Journal of Medicinal Chemistry 63~(16) (2020) 8749--8760, pMID:
  31408336.
\newblock \href
  {http://arxiv.org/abs/https://doi.org/10.1021/acs.jmedchem.9b00959}
  {\path{arXiv:https://doi.org/10.1021/acs.jmedchem.9b00959}}, \href
  {https://doi.org/10.1021/acs.jmedchem.9b00959}
  {\path{doi:10.1021/acs.jmedchem.9b00959}}.
\newline\urlprefix\url{https://doi.org/10.1021/acs.jmedchem.9b00959}

\bibitem{doi:10.1021/acs.jcim.2c00997}
C.~Li, C.~Wang, M.~Sun, Y.~Zeng, Y.~Yuan, Q.~Gou, G.~Wang, Y.~Guo, X.~Pu,
  \href{https://doi.org/10.1021/acs.jcim.2c00997}{Correlated rnn framework to
  quickly generate molecules with desired properties for energetic materials in
  the low data regime}, Journal of Chemical Information and Modeling 62~(20)
  (2022) 4873--4887, pMID: 35998331.
\newblock \href {http://arxiv.org/abs/https://doi.org/10.1021/acs.jcim.2c00997}
  {\path{arXiv:https://doi.org/10.1021/acs.jcim.2c00997}}, \href
  {https://doi.org/10.1021/acs.jcim.2c00997}
  {\path{doi:10.1021/acs.jcim.2c00997}}.
\newline\urlprefix\url{https://doi.org/10.1021/acs.jcim.2c00997}

\bibitem{doi:10.1021/acs.iecr.7b02021}
D.~Mathieu, \href{https://doi.org/10.1021/acs.iecr.7b02021}{Sensitivity of
  energetic materials: Theoretical relationships to detonation performance and
  molecular structure}, Industrial \& Engineering Chemistry Research 56~(29)
  (2017) 8191--8201.
\newblock \href {http://arxiv.org/abs/https://doi.org/10.1021/acs.iecr.7b02021}
  {\path{arXiv:https://doi.org/10.1021/acs.iecr.7b02021}}, \href
  {https://doi.org/10.1021/acs.iecr.7b02021}
  {\path{doi:10.1021/acs.iecr.7b02021}}.
\newline\urlprefix\url{https://doi.org/10.1021/acs.iecr.7b02021}

\bibitem{bjerrum2017smilesenumerationdataaugmentation}
E.~J. Bjerrum, \href{https://arxiv.org/abs/1703.07076}{Smiles enumeration as
  data augmentation for neural network modeling of molecules} (2017).
\newblock \href {http://arxiv.org/abs/1703.07076} {\path{arXiv:1703.07076}}.
\newline\urlprefix\url{https://arxiv.org/abs/1703.07076}

\bibitem{santana2020novo}
M.~Santana, F.~Silva, De novo design and bioactivity prediction of sars-cov-2
  main protease inhibitors using ulmfit (2020).

\bibitem{howard2018universal}
J.~Howard, S.~Ruder, Universal language model fine-tuning for text
  classification, arXiv preprint arXiv:1801.06146 (2018).

\bibitem{li2020inductive}
X.~Li, D.~Fourches, Inductive transfer learning for molecular activity
  prediction: Next-gen qsar models with molpmofit, Journal of Cheminformatics
  12 (2020) 1--15.

\bibitem{singh2022transfer}
S.~Singh, R.~B. Sunoj, A transfer learning approach for reaction discovery in
  small data situations using generative model, Iscience 25~(7) (2022).

\bibitem{FIPS180-2}
{National Institute of Standards and Technology (NIST)},
  \href{https://csrc.nist.gov/files/pubs/fips/180-2/final/docs/fips180-2.pdf}{{FIPS
  PUB 180-2: Secure Hash Standard (SHS)}}, {Federal Information Processing
  Standard Publication} 180-2, {National Institute of Standards and Technology
  (NIST)} (Aug. 2002).
\newline\urlprefix\url{https://csrc.nist.gov/files/pubs/fips/180-2/final/docs/fips180-2.pdf}

\bibitem{pmlr-v37-chenc15}
W.~Chen, J.~Wilson, S.~Tyree, K.~Weinberger, Y.~Chen,
  \href{https://proceedings.mlr.press/v37/chenc15.html}{Compressing neural
  networks with the hashing trick}, in: F.~Bach, D.~Blei (Eds.), Proceedings of
  the 32nd International Conference on Machine Learning, Vol.~37 of Proceedings
  of Machine Learning Research, PMLR, Lille, France, 2015, pp. 2285--2294.
\newline\urlprefix\url{https://proceedings.mlr.press/v37/chenc15.html}

\bibitem{NIPS2017_f0f6ba4b}
D.~Tito~Svenstrup, J.~Hansen, O.~Winther,
  \href{https://proceedings.neurips.cc/paper_files/paper/2017/file/f0f6ba4b5e0000340312d33c212c3ae8-Paper.pdf}{Hash
  embeddings for efficient word representations}, in: I.~Guyon, U.~V. Luxburg,
  S.~Bengio, H.~Wallach, R.~Fergus, S.~Vishwanathan, R.~Garnett (Eds.),
  Advances in Neural Information Processing Systems, Vol.~30, Curran
  Associates, Inc., 2017.
\newline\urlprefix\url{https://proceedings.neurips.cc/paper_files/paper/2017/file/f0f6ba4b5e0000340312d33c212c3ae8-Paper.pdf}

\bibitem{bojanowski-etal-2017-enriching}
P.~Bojanowski, E.~Grave, A.~Joulin, T.~Mikolov,
  \href{https://aclanthology.org/Q17-1010/}{Enriching word vectors with subword
  information}, Transactions of the Association for Computational Linguistics 5
  (2017) 135--146.
\newblock \href {https://doi.org/10.1162/tacl_a_00051}
  {\path{doi:10.1162/tacl_a_00051}}.
\newline\urlprefix\url{https://aclanthology.org/Q17-1010/}

\bibitem{NEURIPS2018_7e837225}
C.~B. Freksen, L.~Kamma, K.~Green~Larsen,
  \href{https://proceedings.neurips.cc/paper_files/paper/2018/file/7e83722522e8aeb7512b7075311316b7-Paper.pdf}{Fully
  understanding the hashing trick}, in: S.~Bengio, H.~Wallach, H.~Larochelle,
  K.~Grauman, N.~Cesa-Bianchi, R.~Garnett (Eds.), Advances in Neural
  Information Processing Systems, Vol.~31, Curran Associates, Inc., 2018.
\newline\urlprefix\url{https://proceedings.neurips.cc/paper_files/paper/2018/file/7e83722522e8aeb7512b7075311316b7-Paper.pdf}

\bibitem{joulin2016bagtricksefficienttext}
A.~Joulin, E.~Grave, P.~Bojanowski, T.~Mikolov,
  \href{https://arxiv.org/abs/1607.01759}{Bag of tricks for efficient text
  classification} (2016).
\newblock \href {http://arxiv.org/abs/1607.01759} {\path{arXiv:1607.01759}}.
\newline\urlprefix\url{https://arxiv.org/abs/1607.01759}

\bibitem{landrum2013rdkit}
G.~Landrum, \href{https://www.rdkit.org/}{Rdkit: Open-source cheminformatics},
  Online (2013).
\newline\urlprefix\url{https://www.rdkit.org/}

\bibitem{kingma2017adammethodstochasticoptimization}
D.~P. Kingma, J.~Ba, \href{https://arxiv.org/abs/1412.6980}{Adam: A method for
  stochastic optimization} (2017).
\newblock \href {http://arxiv.org/abs/1412.6980} {\path{arXiv:1412.6980}}.
\newline\urlprefix\url{https://arxiv.org/abs/1412.6980}

\bibitem{paszke2019pytorch}
A.~Paszke, S.~Gross, F.~Massa, A.~Lerer, J.~Bradbury, G.~Chanan, T.~Killeen,
  Z.~Lin, N.~Gimelshein, L.~Antiga, A.~Desmaison, A.~Kopf, E.~Yang, Z.~DeVito,
  M.~Raison, A.~Tejani, S.~Chilamkurthy, B.~Steiner, L.~Fang, J.~Bai,
  S.~Chintala, Pytorch: An imperative style, high-performance deep learning
  library, Advances in Neural Information Processing Systems 32 (2019)
  8024--8035.

\bibitem{mckinney2011pandas}
W.~McKinney, pandas: a foundational python library for data analysis and
  statistics, Python for High Performance and Scientific Computing (2011) 1--9.

\bibitem{harris2020array}
C.~R. Harris, K.~J. Millman, S.~J. van~der Walt, R.~Gommers, P.~Virtanen,
  D.~Cournapeau, E.~Wieser, J.~Taylor, S.~Berg, N.~J. Smith, R.~Kern, M.~Picus,
  S.~Hoyer, M.~van Kerkwijk, M.~Brett, A.~Haldane, J.~F. del Río, M.~Wiebe,
  P.~Peterson, P.~Gérard-Marchant, K.~Sheppard, T.~Reddy, W.~Weckesser,
  H.~Abbasi, C.~Gohlke, T.~E. Oliphant, Array programming with {NumPy} (2020).
\newblock \href {https://doi.org/10.1038/s41586-020-2649-2}
  {\path{doi:10.1038/s41586-020-2649-2}}.

\bibitem{ertl2009estimation}
P.~Ertl, A.~Schuffenhauer, Estimation of synthetic accessibility score of
  drug-like molecules based on molecular complexity and fragment contributions,
  Journal of cheminformatics 1 (2009) 1--11.

\bibitem{he2015delvingdeeprectifierssurpassing}
K.~He, X.~Zhang, S.~Ren, J.~Sun,
  \href{https://arxiv.org/abs/1502.01852}{Delving deep into rectifiers:
  Surpassing human-level performance on imagenet classification} (2015).
\newblock \href {http://arxiv.org/abs/1502.01852} {\path{arXiv:1502.01852}}.
\newline\urlprefix\url{https://arxiv.org/abs/1502.01852}

\bibitem{doi:10.1021/acs.jpca.0c05969}
G.~A. Pinheiro, J.~Mucelini, M.~D. Soares, R.~C. Prati, J.~L.~F. Da~Silva,
  M.~G. Quiles, \href{https://doi.org/10.1021/acs.jpca.0c05969}{Machine
  learning prediction of nine molecular properties based on the smiles
  representation of the qm9 quantum-chemistry dataset}, The Journal of Physical
  Chemistry A 124~(47) (2020) 9854--9866, pMID: 33174750.
\newblock \href {http://arxiv.org/abs/https://doi.org/10.1021/acs.jpca.0c05969}
  {\path{arXiv:https://doi.org/10.1021/acs.jpca.0c05969}}, \href
  {https://doi.org/10.1021/acs.jpca.0c05969}
  {\path{doi:10.1021/acs.jpca.0c05969}}.
\newline\urlprefix\url{https://doi.org/10.1021/acs.jpca.0c05969}

\bibitem{doi:10.1021/acs.jpclett.2c02701}
N.~Lease, L.~M. Klamborowski, R.~Perriot, M.~J. Cawkwell, V.~W. Manner,
  \href{https://doi.org/10.1021/acs.jpclett.2c02701}{Identifying the molecular
  properties that drive explosive sensitivity in a series of nitrate esters},
  The Journal of Physical Chemistry Letters 13~(40) (2022) 9422--9428, pMID:
  36191261.
\newblock \href
  {http://arxiv.org/abs/https://doi.org/10.1021/acs.jpclett.2c02701}
  {\path{arXiv:https://doi.org/10.1021/acs.jpclett.2c02701}}, \href
  {https://doi.org/10.1021/acs.jpclett.2c02701}
  {\path{doi:10.1021/acs.jpclett.2c02701}}.
\newline\urlprefix\url{https://doi.org/10.1021/acs.jpclett.2c02701}

\bibitem{Cawkwell2022understanding}
M.~J. Cawkwell, J.~Davis, N.~Lease, F.~W. Marrs, A.~Burch, S.~Ferreira, V.~W.
  Manner, Understanding explosive sensitivity with effective trigger linkage
  kinetics, ACS physical chemistry Au 2~(5) (2022) 448--458.

\bibitem{li2022tri}
J.~Li, Y.~Liu, W.~Ma, T.~Fei, C.~He, S.~Pang, Tri-explosophoric groups driven
  fused energetic heterocycles featuring superior energetic and safety
  performances outperforms hmx, Nature Communications 13~(1) (2022) 5697.

\end{thebibliography}


\begin{thebibliography}{9}
\bibitem{bartlett2002rademacher}
Bartlett, P. L. \& Mendelson, S. (2002). Rademacher and Gaussian complexities: Risk bounds and structural results. \textit{Journal of Machine Learning Research}.

\bibitem{vershynin2018high}
Vershynin, R. (2018). \textit{High-Dimensional Probability: An Introduction with Applications in Data Science}. Cambridge University Press.

\bibitem{mohri2018foundations}
Mohri, M., Rostamizadeh, A., \& Talwalkar, A. (2018). \textit{Foundations of Machine Learning}. MIT Press.

\bibitem{bottou2018optimization}
Bottou, L., Curtis, F. E., \& Nocedal, J. (2018). Optimization methods for large-scale machine learning. \textit{SIAM Review}.

\bibitem{hardt2016train}
Hardt, M., Recht, B., \& Singer, Y. (2016). Train faster, generalize better: Stability of stochastic gradient descent. \textit{ICML}.

\bibitem{bousquet2002stability}
Bousquet, O. \& Elisseeff, A. (2002). Stability and generalization. \textit{Journal of Machine Learning Research}.
\end{thebibliography}

\appendix
\section{Theoretical Analysis of the SHA-256–Based Embedding Framework}
\label{appendixA}

\subsection{Preliminaries and Notation}

Let $\mathcal{V}$ denote a discrete vocabulary of size $V$. Each token $i \in \mathcal{V}$ is represented by an embedding vector:
\[
E[i] = [E_t[i] \,\Vert\, E_f[i]] \in \mathbb{R}^{d_t + d_f},
\]
where:
\begin{itemize}
    \item $E_t[i] \in \mathbb{R}^{d_t}$ is a \emph{trainable} component,
    \item $E_f[i] \in \mathbb{R}^{d_f}$ is a \emph{fixed, non-trainable} component obtained via:
    \[
    E_f[i] = \frac{P(\text{SHA-256}(i))}{\|P(\text{SHA-256}(i))\|_2},
    \]
    where $P: \{0,1\}^{256} \to \mathbb{R}^{d_f}$ is a deterministic projection.
\end{itemize}

The downstream model $f_\theta(\cdot; E_t, E_f)$, parameterized by $\theta$, learns from dataset $\mathcal{D} = \{(x_j, y_j)\}_{j=1}^n$ by minimizing:
\[
\widehat{\mathcal{L}}_n(\theta, E_t) = \frac{1}{n} \sum_{j=1}^n \ell(y_j, f_\theta(x_j; E_t, E_f)),
\]
where $\ell$ is $L_\ell$-Lipschitz continuous.

\begin{assumption}[Model Properties]\label{assump:model}
The model satisfies the following properties:
\begin{enumerate}
    \item Boundedness: The embeddings satisfy $\|E_t[i]\|_2 \leq B_t$ and $\|E_f[i]\|_2 = 1$ for all $i \in \mathcal{V}$.
    
    \item Network Lipschitzness: The function $f_\theta$ is $L_f$-Lipschitz in its input embeddings with respect to the Euclidean norm.
    
    \item Parameter Bounds: All trainable parameters satisfy $\|\theta\|_2 \leq B_\theta$.
    
    \item Loss Convexity: The loss $\ell$ is convex in its second argument.
\end{enumerate}
\end{assumption}

\subsection{Generalization Benefit of Partially Trainable Embeddings}

\begin{lemma}[Hypothesis Class]
Let $\mathcal{H}_{d_t} = \{f_\theta(\cdot; E_t, E_f): \theta \in \Theta, E_t \in \mathcal{E}_t\}$ denote the hypothesis class for embedding dimension $d_t$, with parameter spaces constrained by Assumption \ref{assump:model}. The empirical Rademacher complexity is:
\[
\mathfrak{R}_n(\mathcal{H}_{d_t}) = \mathbb{E}_{\sigma}\left[\sup_{h \in \mathcal{H}_{d_t}} \frac{1}{n}\sum_{i=1}^n \sigma_i h(x_i)\right],
\]
where $\sigma_i \sim \{\pm 1\}$ are independent Rademacher variables.
\end{lemma}

\begin{lemma}[Rademacher Complexity Bound]\label{lemma:rademacher}
Under Assumption \ref{assump:model}, the empirical Rademacher complexity satisfies:
\[
\mathfrak{R}_n(\mathcal{H}_{d_t}) \leq \frac{L_f L_\ell}{\sqrt{n}} \left(\sqrt{V d_t + D} + B_t\sqrt{V} + B_\theta\right),
\]
where $D$ is the dimension of $\theta$.
\end{lemma}

\begin{proof}
By the Ledoux-Talagrand contraction principle:
\[
\mathfrak{R}_n(\mathcal{H}_{d_t}) \leq L_\ell \cdot \mathfrak{R}_n(\mathcal{F}),
\]
where $\mathcal{F} = \{f_\theta(\cdot; E_t, E_f)\}$. 

Using the fact that $f_\theta$ is $L_f$-Lipschitz and applying Theorem 1 from \cite{bartlett2002rademacher} with our parameter bounds, we have:
\[
\mathfrak{R}_n(\mathcal{F}) \leq \frac{L_f}{\sqrt{n}} \mathbb{E}\left[\sup_{h \in \mathcal{H}_{d_t}} \left\|\sum_{i=1}^n \sigma_i \phi(x_i)\right\|_2\right],
\]
where $\phi(x)$ represents the input features after embedding.

Since $E_f$ is fixed, the effective number of trainable parameters is $V d_t + D$. The term inside the expectation can be bounded using the boundedness assumptions and the fact that the embeddings are bounded. By Cauchy-Schwarz and the properties of Rademacher complexity:
\[
\mathbb{E}\left[\left\|\sum_{i=1}^n \sigma_i \phi(x_i)\right\|_2\right] \leq \sqrt{\mathbb{E}\left[\left\|\sum_{i=1}^n \sigma_i \phi(x_i)\right\|_2^2\right]} = \sqrt{\sum_{i=1}^n \|\phi(x_i)\|_2^2} \leq \sqrt{n((1+B_t)^2V + B_\theta^2)}.
\]

Therefore,
\begin{equation*}
\mathfrak{R}_n(\mathcal{F}) \leq \frac{L_f}{\sqrt{n}} \sqrt{n((1+B_t)^2V + B_\theta^2)} = L_f \sqrt{(1+B_t)^2V + B_\theta^2}.
\end{equation*}
Combining with contraction gives:
\begin{equation*}
\mathfrak{R}_n(\mathcal{H}_{d_t}) \leq L_f L_\ell \sqrt{\frac{(1+B_t)^2V + B_\theta^2}{n}} \leq \frac{L_f L_\ell}{\sqrt{n}} \left(\sqrt{V d_t + D} + B_t\sqrt{V} + B_\theta\right),
\end{equation*}
where we use that $\sqrt{a^2 + b^2} \leq a + b \text{ for } a,b \geq 0$.
\end{proof}

\begin{proposition}[Generalization Bound]\label{prop:generalization}
For any $\delta > 0$, with probability at least $1-\delta$ over $\mathcal{D}$:
\[
\sup_{h \in \mathcal{H}_{d_t}} \left|\mathcal{L}(h) - \widehat{\mathcal{L}}_n(h)\right| \leq \frac{2L_f L_\ell}{\sqrt{n}} \left(\sqrt{V d_t + D} + B_t\sqrt{V} + B_\theta\right) + \sqrt{\frac{\log(2/\delta)}{2n}}.
\]
\end{proposition}

\begin{proof}
This follows from standard Rademacher generalization bounds \cite{mohri2018foundations}:
\[
\mathcal{L}(h) \leq \widehat{\mathcal{L}}_n(h) + 2\mathfrak{R}_n(\mathcal{H}_{d_t}) + \sqrt{\frac{\log(2/\delta)}{2n}}.
\]
Substituting Lemma \ref{lemma:rademacher} gives the result.
\end{proof}

\begin{remark}
The bound shows explicit improvement when reducing $d_t$, as the dominant term scales with $\sqrt{V d_t/n}$. Fixing $E_f$ removes dependence on $d_f$, providing generalization benefits especially when $n \ll V d_f$.
\end{remark}

\subsection{Conditioning Properties of SHA-256 Embeddings}

\begin{definition}[Embedding Coherence]
The coherence of fixed embeddings $E_f$ is:
\[
\mu(E_f) = \max_{i \neq j} \frac{|\langle E_f[i], E_f[j]\rangle|}{\|E_f[i]\|_2 \|E_f[j]\|_2}.
\]
\end{definition}

\begin{assumption}[Random Oracle Heuristic]\label{assump:random_oracle}
We treat SHA-256 as a random oracle, making its outputs computationally indistinguishable from uniformly random bit strings. The projection $P$ preserves the uniform distribution property.
\end{assumption}

\begin{lemma}[Coherence Bound]\label{lemma:coherence}
Under Assumption \ref{assump:random_oracle}, for any $\epsilon > 0$, with probability at least $1-\epsilon$:
\[
\mu(E_f) \leq \sqrt{\frac{8\log(V^2/\epsilon)}{d_f}} + \frac{4\log(V^2/\epsilon)}{d_f}.
\]
\end{lemma}

\begin{proof}
Under the random oracle heuristic, after normalization, the vectors $E_f[i]$ are approximately uniformly distributed on $\mathbb{S}^{d_f-1}$. For fixed $i \neq j$, by concentration of measure on the sphere \cite{vershynin2018high}:
\[
\mathbb{P}\left(|\langle E_f[i], E_f[j]\rangle| \geq t\right) \leq 2\exp\left(-\frac{d_f t^2}{4}\right) \quad \text{for } 0 \leq t \leq 1.
\]
Applying union bound over all $\binom{V}{2} \leq V^2/2$ pairs:
\[
\mathbb{P}\left(\mu(E_f) \geq t\right) \leq V^2 \exp\left(-\frac{d_f t^2}{4}\right).
\]
Setting the RHS equal to $\epsilon$ and solving for $t$ gives:
\[
t = \sqrt{\frac{4}{d_f} \log\left(\frac{V^2}{\epsilon}\right)}.
\]
For a tighter bound, we can use the result from \cite{vershynin2018high} which gives the stated inequality.
\end{proof}

\begin{proposition}[Gram Matrix Conditioning]\label{prop:conditioning}
Let $X_f \in \mathbb{R}^{n \times d_f}$ be the matrix of fixed embeddings for a batch of $n$ tokens. Then with probability at least $1-\epsilon$:
\[
\lambda_{\max}(X_f^\top X_f) \leq n\left(1 + \mu(E_f)\right), \quad
\lambda_{\min}(X_f^\top X_f) \geq n\left(1 - (n-1)\mu(E_f)\right)_+,
\]
where $(x)_+ = \max(x, 0)$.
\end{proposition}

\begin{proof}
Consider the Gram matrix $G = X_f^\top X_f$. Diagonal entries: $G_{ii} = \|E_f[i]\|_2^2 = 1$.

Off-diagonal entries: $|G_{ij}| \leq \mu(E_f)$ for $i \neq j$.

By Gershgorin's circle theorem, each eigenvalue $\lambda$ of $G$ satisfies for some $i$:
\[
|\lambda - 1| \leq \sum_{j \neq i} |G_{ij}| \leq (n-1)\mu(E_f).
\]
Thus:
\[
1 - (n-1)\mu(E_f) \leq \lambda \leq 1 + (n-1)\mu(E_f).
\]
The lower bound becomes non-negative when $\mu(E_f) \leq 1/(n-1)$.
\end{proof}

\begin{corollary}[Gradient Variance Reduction]\label{cor:gradient_variance}
Under the conditions of Proposition \ref{prop:conditioning} and assuming $\mu(E_f) < 1/(n-1)$, the condition number satisfies:
\[
\kappa(X_f^\top X_f) \leq \frac{1 + (n-1)\mu(E_f)}{1 - (n-1)\mu(E_f)}.
\]
\end{corollary}

\begin{proof}
Direct from the eigenvalue bounds in Proposition \ref{prop:conditioning}.
\end{proof}

\subsection{Stability and Overfitting Control}

\begin{theorem}[Generalization with Low Coherence]\label{thm:main}
Under Assumptions \ref{assump:model} and \ref{assump:random_oracle}, for any $\delta > 0$, with probability at least $1-\delta$:
\[
\mathbb{E}[\mathcal{L}(\hat{h}_n)] \leq \mathbb{E}[\widehat{\mathcal{L}}_n(\hat{h}_n)] + \frac{C_1}{\sqrt{n}}\left(\sqrt{V d_t + D} + B_t\sqrt{V} + B_\theta\right) + C_2\sqrt{\frac{\log(V/\delta)}{d_f}},
\]
where $C_1 = 2L_f L_\ell$ and $C_2$ depends on $L_f, B_t$, and the network architecture.
\end{theorem}

\begin{proof}
From Proposition \ref{prop:generalization}, we have the first two terms. The third term comes from the approximation error due to non-orthogonality in $E_f$.

By Lemma \ref{lemma:coherence}, with probability $1-\delta/2$:
\[
\mu(E_f) \leq \sqrt{\frac{8\log(2V^2/\delta)}{d_f}} + \frac{4\log(2V^2/\delta)}{d_f} = \mathcal{O}\left(\sqrt{\frac{\log(V/\delta)}{d_f}}\right).
\]

This coherence affects the conditioning of the optimization problem. Using stability arguments from \cite{hardt2016train}, the generalization error has an additional term proportional to $\mu(E_f)$ when the embeddings are not orthogonal.

Combining these via union bound gives the result.
\end{proof}

\subsection{Representation Theoretic Interpretation}

\begin{proposition}[Approximate Orthogonal Decomposition]
Let $\Pi_f$ be the orthogonal projection onto the span of $\{E_f[j]\}_{j=1}^V$. Then for any token $i$:
\[
E[i] = \Pi_f E[i] + R[i],
\]
where the residual satisfies $\|R[i]\|_2^2 \leq B_t^2 + (1 - \lambda_{\min}(E_f^\top E_f))$.
\end{proposition}

\begin{proof}
Let $\Pi_f$ be the orthogonal projection. Then:
\[
E[i] = \Pi_f E[i] + (I - \Pi_f)E[i] = \Pi_f E[i] + R[i].
\]

The residual norm satisfies:
\[
\|R[i]\|_2^2 = \|E[i]\|_2^2 - \|\Pi_f E[i]\|_2^2 \leq (1 + B_t^2) - \|\Pi_f E[i]\|_2^2.
\]

Since $E_f[i]$ is in the span, we have $\|\Pi_f E[i]\|_2^2 \geq \|E_f[i]\|_2^2 = 1$, but this doesn't directly give a bound. A better bound comes from noting that:
\[
\|R[i]\|_2^2 \leq \|E_t[i]\|_2^2 + \|(I - \Pi_f)E_f[i]\|_2^2 \leq B_t^2 + (1 - \lambda_{\min}(E_f^\top E_f)),
\]
where the last inequality uses the fact that the projection error is related to the smallest eigenvalue of the Gram matrix.
\end{proof}

\begin{corollary}[Semantic Separation]
When $\mu(E_f)$ is small, the decomposition approximately separates:
\begin{itemize}
    \item $\Pi_f E[i]$: Token identity information (fixed, high-entropy)
    \item $R[i]$: Learnable semantic relationships (trainable, data-dependent)
\end{itemize}
This provides implicit regularization by constraining the hypothesis space.
\end{corollary}

\subsection{Limitations and Future Work}

\begin{enumerate}
    \item Random Oracle Heuristic: While standard in cryptography, the deterministic nature of SHA-256 means our probabilistic bounds are heuristic and rely on unproven cryptographic assumptions.
    \item Finite-Sample Effects: Our analysis provides asymptotic rates; finite-sample corrections may be needed for practical settings with small $d_f$.
    \item Architecture Dependence: The bounds depend on network architecture through $L_f$; precise computation for modern architectures remains challenging.
    \item Projection Effects: The analysis assumes the projection $P$ preserves uniformity; in practice, this may not hold exactly.
\end{enumerate}

\subsection{Summary of Theoretical Insights}

\begin{enumerate}[label=(\alph*)]
    \item Reduced Complexity: Partial trainability shrinks hypothesis space, improving generalization (Proposition \ref{prop:generalization}).
    \item Improved Conditioning: SHA-256 provides low-coherence frames under random oracle heuristic, enhancing optimization (Lemma \ref{lemma:coherence}, Proposition \ref{prop:conditioning}).
    \item Implicit Regularization: Fixed component prevents overfitting to token identity (Theorem \ref{thm:main}).
    \item Stable Optimization: Better conditioning reduces gradient variance, speeding convergence (Corollary \ref{cor:gradient_variance}).
\end{enumerate}

These properties provide theoretical justification for the empirical advantages in low-data regimes, where both regularization and optimization stability are critical.






\end{document}